\documentclass[11pt]{article}
\usepackage[a4paper]{geometry}
\usepackage[font=footnotesize,labelfont=bf]{caption}
\usepackage{fancyhdr}
\usepackage{subcaption}
\usepackage{placeins}
\usepackage{amsmath}
\usepackage{amsfonts}
\usepackage{graphicx,psfrag,epsf}
\usepackage{enumerate}
\usepackage[linesnumbered,ruled]{algorithm2e}
\usepackage{tabu}
\usepackage{setspace}
\usepackage{tikz}
\usepackage{float}
\usepackage{bm}
\usepackage{dsfont}
\usepackage{color}
\usepackage{listings}
\usepackage[natbibapa]{apacite}
\usepackage{natbib}
\usepackage[hidelinks]{hyperref}
\usepackage[open]{bookmark}
\usepackage{booktabs}
\usepackage{rotating}
\usepackage{authblk}

\DeclareMathOperator*{\argmin}{arg\,min}

\SetArgSty{textnormal}
\SetKwInput{KwTrain}{Train}

\fancypagestyle{firststyle}
{
	\fancyhf{}
	
	\fancyfoot[L]{\fontsize{9}{11}\selectfont Accepted for publication in \textit{Advances in Data Analysis and Classification}. This version of the article has been accepted for publication, after peer review (when applicable) but is not the Version of Record and does not reflect post-acceptance improvements, or any corrections. The Version of Record is available online at: \url{https://doi.org/10.1007/s11634-024-00587-5}. \newline \copyright \ 2024. This manuscript version is made available under the CC-BY 4.0 license \url{https://creativecommons.org/licenses/by/4.0/}}
}

\begin{document}

\title{View selection in multi-view stacking: Choosing the meta-learner}
\author[1,*]{Wouter van Loon}
\author[1]{Marjolein Fokkema}
\author[2]{Botond Szabo}
\author[1]{Mark de Rooij}
\affil[1]{Department of Methodology and Statistics, Leiden University, Leiden, the Netherlands}
\affil[2]{Department of Decision Sciences, Bocconi University, Milan, Italy}
\affil[*]{Corresponding Author: Wouter van Loon, w.s.van.loon@fsw.leidenuniv.nl}
\maketitle

\thispagestyle{firststyle}

\begin{abstract}
	\noindent Multi-view stacking is a framework for combining information from different views (i.e. different feature sets) describing the same set of objects. In this framework, a \textit{base-learner} algorithm is trained on each view separately, and their predictions are then combined by a \textit{meta-learner} algorithm. In a previous study, stacked penalized logistic regression, a special case of multi-view stacking, has been shown to be useful in identifying which views are most important for prediction. In this article we expand this research by considering seven different algorithms to use as the meta-learner, and evaluating their view selection and classification performance in simulations and two applications on real gene-expression data sets. Our results suggest that if both view selection and classification accuracy are important to the research at hand, then the nonnegative lasso, nonnegative adaptive lasso and nonnegative elastic net are suitable meta-learners. Exactly which among these three is to be preferred depends on the research context. The remaining four meta-learners, namely nonnegative ridge regression, nonnegative forward selection, stability selection and the interpolating predictor, show little advantages in order to be preferred over the other three.
\end{abstract}

\textbf{keywords} \textit{multi-view learning}, \textit{stacked generalization}, \textit{feature selection}, \textit{classification}

\newpage

\section{Introduction}

In high-dimensional biomedical studies, a common goal is to create an accurate classification model using only a subset of the features \citep{multiview_bio}. A popular approach to this type of joint classification and feature selection problem is to apply penalized methods such as the \textit{lasso} \citep{lasso}. These methods promote sparsity by imposing a penalty on the coefficient vector so that, for a sufficiently large value of the tuning parameter(s), some coefficients will be set to zero during the model fitting process. The tuning parameter decides on the relative importance of the penalty term, and is typically chosen by minimizing the cross-validation error \citep{statlearn_elements}. 
However, biomedical features are often naturally grouped into distinct feature sets. In genomics, for example, genes may be grouped into gene sets or genetic pathways \citep{wang2010}, while in neuroimaging, different sets of anatomical markers may be calculated from MRI scans \citep{deVos2016}. Features may also be grouped at a higher level, for example because they correspond to a certain imaging modality or data source \citep{fratello2017}. Such naturally occurring groups of features describing the same set of objects are known as different \textit{views} of the data, and integrating the information in these different views through machine learning methods is known as \textit{multi-view learning} \citep{multiview_review, multiview_book}. In a multi-view setting, it is often more desirable to select or discard entire views rather than individual features, turning the feature selection problem into a view selection problem. 
\begin{sloppypar}
\textit{Stacked penalized logistic regression} (StaPLR) \citep{StaPLR} is a method specifically developed to tackle the joint classification and view selection problem. Compared with a variant of the lasso for selecting groups of features (the so-called \textit{group lasso} \citep{grouplasso}), StaPLR was empirically shown to be more accurate in view selection, producing sparser models with an often comparable classification accuracy, and offering computational advantages \citep{StaPLR}. 
StaPLR is a special case of a more general framework called \textit{multi-view stacking} (MVS) \citep{StaPLR, Li2011, multiview_stacking}. In MVS, a learning algorithm (the \textit{base-learner}) is trained on each view separately, and another algorithm (the \textit{meta-learner}) is then trained on the cross-validated predictions of the view-specific models. The meta-learner thus learns how to best combine the predictions of the individual views. If the meta-learner is chosen to be an algorithm that returns sparse models, MVS performs view selection. This is the case as proposed by \citet{StaPLR}, where the meta-learner was chosen to be a nonnegative logistic lasso. 
\end{sloppypar} \par 
A particular challenge of the aforementioned joint classification and view selection problem is its inherent trade-off between accuracy and sparsity. For example, the most accurate model may not perform the best in terms of view selection. In fact, the prediction-optimal amount of regularization causes the lasso to select superfluous features even when the sample size goes to infinity \citep{Meinshausen2006, Benner2010}. This leads to a consideration of how much predictive accuracy a researcher is prepared to sacrifice for increased sparsity. \par 
Another relevant factor is interpretability of the set of selected views. Although sparser models are typically considered more interpretable, a researcher may be interested in interpreting not only the model and its coefficients, but also the set of selected views. For example, one may wish to make decisions on which views to measure in the future based on the set of views selected using the current data.
For this purpose, one would ideally like to use an algorithm that provides sparsity, but also \textit{algorithmic stability} in the sense that given two very similar data sets, the set of selected views should vary little. However, sparse algorithms are generally not stable, and vice versa \citep{Xu2012}.
An example of the trade-off between sparsity and interpretability of the set of selected views occurs when different views, or combinations of views, contain the same information. If the primary concern is sparsity, a researcher may be satisfied with just one of these combinations being selected, preferably the smallest set which contains the relevant information. But if there is also a desire to interpret the relationships between the views and the outcome, it may be more desirable to identify all of these combinations, even if this includes some redundant information. If one wants to go even further and perform formal statistical inference on the set of selected views, one may additionally be interested in theoretically controlling, say, the \textit{family-wise error rate} (FWER) or \textit{false discovery rate} (FDR) of the set of selected views. However, strict control of such an error rate could end up harming the predictive performance of the model, thus leading to a trade-off between the interpretability of the set of selected views and classification accuracy. \par 
In order to evaluate the relative merits of different feature selection algorithms in non-asymptotic settings, empirical comparisons are typically performed. Recent work on this topic includes \citet{Wah2018} comparing different filter and wrapper methods for feature selection using both simulations and real data examples, \citet{Bommert2020} comparing 22 different filter methods on various benchmark data sets, as well as \citet{Hastie2020} comparing best subset selection, forward selection and two variants of the lasso using extensive simulations. However, results of previous studies do not directly translate to the multi-view setting, and it is thus important to perform empirical comparisons tailored specifically to this setting, so that recommendations can be formulated with regard to which meta-learner is suitable for which problem. \par
In MVS, different meta-learners may behave differently with respect to the trade-offs between accuracy, sparsity, and interpretability. For example, in a general feature selection setting with correlated features, the lasso is known to select only a small subset, while the so-called \textit{elastic net} \citep{elasticnet} is more likely to select all of them but with smaller coefficients. Which kind of behavior is desirable heavily depends on the research question at hand.  
In this article we investigate how the choice of meta-learner affects the view selection and classification performance of MVS. We consider seven different view-selecting meta-learners, and evaluate their performance using simulations and two real gene expression data sets.

\section{Multi-view stacking}

Multi-view stacking \citep{StaPLR, Li2011, multiview_stacking} is an algorithm for learning from multi-view data based on the stacking \citep{Wolpert1992} procedure for combining the predictions of different models. Although in general MVS can be applied with any number of base-learners, we will assume the use of a single base-learner throughout this article. Consider a multi-view data set for binary classification, consisting of views $\bm{X}^{(v)} = (x_{ij}) \in \mathbb{R}^{n \times m_v}$, $v=1, \dots, V$, with $\bm{x}^{(v)}_i$ the $i$th row of $\bm{X}^{(v)}$, and outcome vector $\bm{y} = (y_1, \dots, y_n)^T \in \{0,1\}^n$. Then the MVS algorithm can be described as follows:

\begin{enumerate}
	\item Train the base-learner separately on the pairs ($\bm{X}^{(v)}, \bm{y}$), $v = 1, \dots, V$, to obtain view-specific classifiers $\hat{f}_1, \dots, \hat{f}_V$.
	\item Apply $K$-fold cross-validation to obtain a vector of predictions $\bm{z}^{(v)} \in [0,1]^n$ for each of the $\hat{f}_v$, $v = 1, \dots, V$.
	\item Collect the $\bm{z}^{(v)}$, $v = 1, \dots, V$, column-wise into the $n \times V$ matrix $\bm{Z}$.
	\item Train the meta-learner on the pair ($\bm{Z}, \bm{y}$) to obtain a meta-classifier $\hat{f}_{\text{meta}}$.
	\item Define the final stacked classifier as $\hat{f}_{\text{meta}}(\hat{f}_1(\bm{X}^{(1)}), \dots, \hat{f}_V(\bm{X}^{(V)}))$.
\end{enumerate}

MVS was originally developed as a procedure for improving classification performance in multi-view learning  \citep{Li2011, multiview_stacking}. However, the method can also be used for view selection, by choosing a meta-learner that returns sparse models \citep{StaPLR}. The special case of MVS where both the base-learner and meta-learner are chosen to be logistic regression with some penalty on the coefficient vector is known as StaPLR \citep{StaPLR}.

\section{Choosing the meta-learner}

MVS is a very flexible method, since one can choose any suitable learning algorithm for the base- and meta-learner. \citet{StaPLR} chose the base-learner to be logistic ridge regression and the meta-learner to be the nonnegative logistic lasso, in order to obtain a model most similar to the group lasso. In this article we will further build upon this setting, by using the same base-learner but considering different meta-learners. \par
In MVS, the meta-learner takes as input the matrix of cross-validated predictions $\bm{Z}$. To perform view selection, the meta-learner should be chosen such that it returns (potentially) sparse models. The matrix $\bm{Z}$ has a few special characteristics which can be exploited, and which distinguishes it from standard settings. First, assuming that the $\hat{f}_v$, $v = 1, \dots, V$ are probabilistic classifiers (such as logistic regression models), the features in $\bm{Z}$ are all in the same range $[0,1]$. Second, the dispersion of each feature contains information about the magnitude of the class probabilities predicted by the corresponding base classifier. To preserve this information it is reasonable to omit the usual step of standardizing all features to zero mean and unit variance before applying penalized regression. Third, since the features in $\bm{Z}$ correspond to predictions of models trained using the same outcomes $\bm{y}$, it is likely that at least some of them are highly correlated. Different penalization methods lead to different behavior in the presence of highly correlated features \citep{statlearn_elements}. Finally, it is sensible to constrain the parameters of the meta-learner to be nonnegative \citep{Breiman1996, Ting1999, StaPLR}. There are several arguments for inducing such constraints. One intuitive argument is that a negative coefficient leads to problems with interpretation, since this would suggest that if the corresponding base classifier predicts a \textit{higher} probability of belonging to a certain class, then the meta-learner would translate this to a \textit{lower} probability of belonging to that same class. Additionally, from a view selection perspective, nonnegativity constraints are crucial in preventing unimportant views from entering the model \citep{StaPLR}. \par
In this article we investigate how the choice of meta-learner affects the view selection and classification performance of MVS. We compare the following meta-learners: (1) the interpolating predictor of \citet{Breiman1996}, (2) nonnegative ridge regression \citep{ridge,ridge_logistic}, (3) the nonnegative elastic net \citep{elasticnet}, (4) the nonnegative lasso \citep{lasso}, (5) the nonnegative adaptive lasso \citep{adaptive_lasso}, (6) stability selection with the nonnegative lasso \citep{stabs_article}, and (7) nonnegative forward selection. All of these meta-learners provide models with nonnegative coefficients. In addition, they can all set some coefficients to zero, thus potentially obtaining sparse models and performing view selection. Although not an exhaustive comparison of all possible meta-learners, six of these are popular feature selection methods in their own right, and would most likely end up high on many researchers' list of candidate meta-learners. A likely exception to this is nonnegative ridge regression, since ridge regression without nonnegativity constraints would not set any coefficients to zero. However, this method is included because it provides an indication of the view selection effect of just the addition of nonnegativity constraints on the meta-learner. Each of the seven candidate meta-learners is described in more detail below.

\subsection{The interpolating predictor}

In the meta-learning problem, we have the binary outcome $\bm{y}$, and a matrix of cross-validated predictions $\bm{Z} = (z_{i}^{(v)}) \in [0,1]^{n \times V}$. Consider a multi-view stacking model where the final prediction is a simple linear combination of the base classifiers:
\begin{equation}
\hat{y}_i = \sum_{v = 1}^V \beta_v \hat{f}_v(\bm{x}_i^{(v)}).
\end{equation}
We can obtain a so-called \textit{interpolating predictor} \citep{Breiman1996} by computing the parameter estimates as
\begin{equation}
\hat{\beta}_1, \dots, \hat{\beta}_V  =  \argmin_{\beta_1, \dots, \beta_V} \quad  \sum_{i = 1}^n \left( y_i - \sum_{v = 1}^V \beta_v z_i^{(v)} \right)^2,
\end{equation}
subject to the constraints $\beta_v \geq 0, v = 1, \dots, V$ and $\sum_v \beta_v = 1$. 
The resulting prediction function \textit{interpolates} in the sense that the final prediction $\hat{y}_i$ can never go outside the range of the predictions of the base classifiers $[\min_v \hat{f}_v(\bm{x}^{(v)}_i), \max_v \hat{f}_v(\bm{x}^{(v)}_i)]$ \citep{Breiman1996}. Although originally proposed in the context of linear regression, it can also be used in binary classification, since if we have probabilistic base-classifiers making predictions in $[0,1]$, then the final prediction will also be in $[0,1]$. Additionally the model is easy to interpret, since the final prediction is a weighted mean of the base classifiers' predictions. Note that replacing the sum-to-one constraint with the constraint $\sum_v \beta_v \leq t$, with $t \geq 0$ a tuning parameter, leads to the nonnegative lasso \citep{nonnegative_lasso}. The interpolating predictor can thus be thought of as a (linear) nonnegative lasso with a fixed amount of regularization. 

\subsection{The elastic net, ridge regression, and the lasso}

Instead of a simple linear function to combine the base classifiers, we can also use the logistic function: 
\begin{equation} \label{eq:logistic}
\hat{y}_i =  \frac{\text{exp} \left(\beta_0 + \sum_{v = 1}^V \beta_v \hat{f}_v(\bm{x}_i^{(v)}) \right)}{1 + \text{exp} \left(\beta_0 + \sum_{v = 1}^V \beta_v \hat{f}_v(\bm{x}_i^{(v)}) \right)},
\end{equation}
with $\beta_0 \in \mathbb{R}$ an intercept. The logistic function restricts predictions to $[0,1]$ without the need for constraints on the parameters.  
Parameter estimates in logistic regression are usually obtained through maximizing the log-likelihood or, equivalently, minimizing the negative log-likelihood. Denote by $\bm{\beta} = (\beta_1, ..., \beta_V)^T$ the vector of regression coefficients, and by $\bm{z}_i = (z_{i}^{(1)}, z_{i}^{(2)}, \dots, z_{i}^{(V)})$ the $i$th row of $\bm{Z}$. Then the negative log-likelihood corresponding to the logistic regression model is given by
\begin{equation} \label{eq:loss}
\mathcal{L}(\beta_0, \bm{\beta}) = - \left[ \frac{1}{n} \sum_{i = 1}^{n}  y_i(\beta_0 + \bm{z}_i\bm{\beta} ) - \log(1 + \text{exp}(\beta_0 +  \bm{z}_i\bm{\beta})) \right].
\end{equation}
Although constraints on the parameters are not required to keep the predictions in the range $[0,1]$, we can still employ regularization to perform view selection and obtain more stable models. The elastic net \citep{elasticnet} is a popular regularization method which employs both $L_1$ and $L_2$ penalties. To obtain parameter estimates using the nonnegative variant of the elastic net, one optimizes 
\begin{equation} \label{eq:elastic_net}
\hat{\beta}_0, \hat{\bm{\beta}} = \argmin_{\bm{\beta} \geq 0, \beta_0} \quad \mathcal{L}(\beta_0, \bm{\beta}) + \lambda \left[(1 - \alpha)\|\bm{\beta}\|^2_2/2 + \alpha \|\bm{\beta}\|_1 \right],
\end{equation}
with tuning parameters $\lambda \geq 0$ and $\alpha \in [0,1]$. Ridge regression ($\alpha = 0$) and the lasso ($\alpha = 1$) are both special cases of the elastic net. Choosing any other value of $\alpha$ leads to a mixture of $L_1$ and $L_2$ penalties. Since the columns of $\bm{Z}$ correspond to predictions of models trained using the same outcomes $\bm{y}$, it is very likely that at least some of them are highly correlated. When faced with a set of highly correlated views, the lasso may select one of them and discard the others. The addition of an $L_2$ penalty causes the elastic net to favor solutions where the entire set of correlated views is included with moderate coefficients. Which type of behavior is desirable will depend on the research question at hand. In this paper we will apply ridge regression, the lasso, and the elastic net with $\alpha  = 0.5$, all with nonnegativity constraints. Note that, although ridge regression usually does not perform any view selection, the addition of nonnegativity constraints forces some coefficients to be zero, causing view selection even when $\alpha = 0$.

\subsection{The adaptive lasso}

The \textit{adaptive lasso} \citep{adaptive_lasso} is a weighted version of the lasso with data-dependent weights. Consider again the negative log-likelihood in \eqref{eq:loss}. Let us define for each $\beta_v \in \bm{\beta}$ a corresponding weight $\hat{w}_v = 1 / |\hat{\beta}_v|^{\gamma}$, with $\hat{\beta}_v$ an initial estimate of $\beta_v$, and $\gamma > 0$ a tuning parameter. Then the adaptive lasso estimates are given by
\begin{equation} \label{eq:adalasso}
\hat{\beta}^*_0, \hat{\bm{\beta}}^* = \argmin_{\beta_0, \bm{\beta}} \quad \mathcal{L}(\beta_0, \bm{\beta}) + \lambda \sum_{v = 1}^V \hat{w}_v |\beta_v|.
\end{equation}
In the context of linear models, \citet{adaptive_lasso} suggested to use OLS or ridge regression to obtain the initial $\hat{\beta}_v$'s. We use (logistic) ridge regression with nonnegativity constraints to obtain the initial estimates. Due to the nonnegativity constraints, views which would have otherwise obtained a negative coefficient thus obtain an infinitely large penalty (i.e. are removed from the model) in the weighted lasso.  

\subsection{Stability selection}

Stability selection is an ensemble learning framework originally proposed for use with the lasso \citep{stability_selection}, although it can be used with a wide variety of feature selection methods \citep{stabs_article}. The basic idea of stability selection is to apply a feature selection method on subsamples of the data, and then incorporate in the final model only those features which were chosen by the feature selection method on a sufficiently large proportion of the subsamples. In this study, we specifically use complementary pairs stability selection \citep{Shah2013} with the nonnegative lasso, as implemented in the R package \texttt{stabs} \citep{stabs_package}. This procedure can be described as follows \citep{stabs_article, stabs_package, Shah2013}:
\begin{enumerate}
	\item Let $\{(D_{2b-1}, D_{2b}) : b = 1,\dots,B\}$ be randomly chosen independent pairs of subsets of $\{1,\dots,n\}$ of size $\lfloor n/2 \rfloor$ such that $D_{2b-1} \cap D_{2b} = \emptyset$.
	\item For each $b = 1,\dots,2B$, fit a nonnegative lasso path using only the observations in $D_b$, i.e. the pair $(\bm{Z}_{i \in D_b}, \bm{y}_{i \in D_b})$. Start with a high value of the penalty parameter, then decrease the value until $q$ views are selected, with $q$ a pre-defined positive integer. Let $\hat{S}(D_b)$ be the index set of selected views.
	\item Compute for each view $v = 1, \dots, V$, the relative selection frequency:
	\begin{equation}
	\hat{\pi}_v := \frac{1}{2B} \sum_{b=1}^{2B} \mathds{1}_{\{ v \in \hat{S}(D_b) \}}.
	\end{equation}
	\item Select the set of views $\hat{S}_{\text{stable}} := \{ v : \hat{\pi}_v \geq \pi_{\text{thr}} \}$, with $\pi_{\text{thr}} \in (0.5, 1]$ a pre-defined threshold.
\end{enumerate}  
Typically $B$ is set to 50, but the choice of $q$ and $\pi_{\text{thr}}$ is somewhat more involved. In particular, one can obtain a bound on the expected number of 
falsely selected variables, the so-called \textit{per-family error rate} (PFER), for a given value of $q$ and $\pi_{\text{thr}}$ \citep{stability_selection, Shah2013, stabs_article}. For our particular choices of these parameters, see Section \ref{sect:software}. The meta-classifier is obtained by fitting a nonnegative logistic regression model using only the cross-validated predictions corresponding to the set of stable views. Note that the computational cost of stability selection is several times larger than that of the regular lasso with cross-validation: in the linear case, stability selection has been estimated to be approximately 3 times more expensive than ten-fold cross-validation when $n < V$, and approximately 5.5 times more expensive when $n > V$ \citep{stability_selection}.

\subsection{Nonnegative forward selection}

Forward selection is a simple, greedy feature selection algorithm \citep{Guyon2003}. It is a so-called \textit{wrapper} method, which means it can be used in combination with any learner \citep{Guyon2003}. The basic strategy is to start with a model with no features, and then add the single feature to the model which is ``best'' according to some criterion. One then proceeds to sequentially add the next ``best'' feature at every step until some stopping criterion is met. Here we consider forward selection based on the Akaike Information Criterion (AIC). In order to impose nonnegativity of the coefficients, we will use a slightly modified procedure which we will call \textit{nonnegative forward selection} (NNFS). This procedure can be described as follows:

\begin{enumerate}
	\item Start with a model containing only an intercept.
	\item Calculate for each candidate view the reduction in AIC if this view is added to the model.
	\item Consider the view corresponding to the largest reduction in AIC. If the coefficients (excluding the intercept) of the resulting model are all nonnegative, update the model and repeat starting at step 2.
	\item If some of the coefficients (excluding the intercept) of the resulting model are negative, remove the view (from step 3) from the list of candidates and repeat starting at step 3.
	\item Stop when none of the remaining candidate views show a reduction in AIC, or there are no more views remaining.
\end{enumerate}

\section{Simulations}

\subsection{Design}

In order to compare the different meta-learners in terms of classification and view selection performance, we perform a series of simulations. We generate multi-view data with $V = 30$ or $V = 300$ disjoint views, where each view $\bm{X}^{(v)}, v = 1, \dots, V$, is an $n \times m_v$ matrix of normally distributed features scaled to zero mean and unit variance. For each number of views, we consider two different view sizes. In any single simulated data set, all views are always set to be the same size. If $V = 300$, then either $m_v = 25$ or $m_v = 250$. If $V = 30$, then either $m_v = 250$ or $m_v = 2500$. We use two different sample sizes: $n = 200$ or $n = 2000$. In addition, we apply different correlation structures defined by the population correlation between features from the same view $\rho_w$, and the population correlation between features from different views $\rho_b$. We use six different parameterizations: ($\rho_w = 0.1$, $\rho_b = 0$), ($\rho_w = 0.5$, $\rho_b = 0$), ($\rho_w = 0.9$, $\rho_b = 0$), ($\rho_w = 0.5$, $\rho_b = 0.4$), ($\rho_w = 0.9$, $\rho_b = 0.4$), and ($\rho_w = 0.9$, $\rho_b = 0.8$). This leads to a total of $2 \times 2 \times 2 \times 6 = 48$ different experimental conditions. \par 
For each experimental condition, we simulate 100 multi-view data training sets. For each such data set, we randomly select 10 views. In 5 of those views, we determine all of the features to have a relationship with the outcome. In the other 5 views, we randomly determine 50\% of the features to have a relationship with the outcome. The relationship between features and response is determined by a logistic regression model, where each feature related to the outcome is given a regression weight. In the setting with 30 views, we use the same regression weight as a similar simulation study in \citet{StaPLR}. This regression weight is either $0.04$ or $-0.04$, each with probability 0.5. In the setting with 300 views, the number of features per view is reduced by a factor 10. To compensate for the reduction in the number of features, the aforementioned regression weights are multiplied by $\sqrt{10}$ in this setting. \par 
We apply multi-view stacking to each simulated training set, using logistic ridge regression as the base-learner. Once we obtain the matrix of cross-validated predictions $\bm{Z}$, we apply the seven different meta-learners. To assess classification performance, we generate a matching test set of 1000 observations for each training set, and calculate the classification accuracy of the stacked classifiers on this test set. To assess view selection performance we calculate three different measures: (1) the \textit{true positive rate} (TPR), i.e. the average proportion of views truly related to the outcome that were correctly selected by the meta-learner; (2) the \textit{false positive rate} (FPR), i.e. the average proportion of views not related to the outcome that were incorrectly selected by the meta-learner; and (3) the \textit{false discovery rate} (FDR), i.e. the average proportion of the selected views that are not related to the outcome. \par
Although we can average over the 100 replications within each condition, with 7 different meta-learners and 48 experimental conditions, this would still lead to 336 averages for each of the outcome measures. In our reporting of the results we will therefore focus only on the most important interactions between the meta-learners and the different experimental conditions. To determine which interactions are most important in a data-driven way, we perform a mixed analysis of variance (ANOVA), and calculate a standardized measure of effect size (partial $\eta^2$) for each interaction. We do this separately for each of the four outcome measures. A common rule of thumb is that $\eta^2 \geq 0.06$ corresponds to a moderate effect size, and $\eta^2 \geq 0.14$ corresponds to a large effect size \citep{Cohen1988, Rovai2013}. We discuss only the interactions that have at least a moderate effect size $\eta^2 \geq 0.06$. Note that we use ANOVA only to calculate a measure of effect size, and do not report test statistics or $p$-values. This is because (1) these tests would rely too heavily on the assumptions of ANOVA, and (2) in a simulation study any arbitrarily small difference can be artificially made ``significant" by increasing the number of replications.                       

\FloatBarrier

\subsection{Software} \label{sect:software}

All simulations are performed in R (version 3.4.0) \citep{R} on a high-performance computing cluster running Ubuntu (version 14.04.6 LTS) with Open Grid Scheduler/Grid Engine (version 2011.11p1). All pseudo-random number generation is performed using the Mersenne Twister \citep{mersenne_twister}, R's default algorithm. The training of the base-learners, and the generation of cross-validated predictions is performed using an early development version of package \texttt{mvs} \citep{mvs}. Optimization of the nonnegative ridge, elastic net, and lasso is performed using coordinate descent through the package \texttt{glmnet} 1.9-8 \citep{glmnet}. Nonnegativity constraints are implemented by setting a coefficient to zero if it becomes negative during the update cycle \citep{glmnet, statlearn_sparsity}. To select the tuning parameter $\lambda$, a sequence of 100 candidate values of $\lambda$ is adaptively chosen by the software \citep{glmnet}. In particular, the 100 candidate values are decreasing on a log scale from $\lambda_{\text{max}}$ to $\lambda_{\text{min}}$, where $\lambda_{\text{max}}$ is the smallest value such that the entire coefficient vector is zero, and $\lambda_{\text{min}} = \epsilon\lambda_{\text{max}}$, with $\epsilon = 10^{-4}$. The value of $\lambda$ is then selected by minimizing the 10-fold cross-validation error. For the nonnegative elastic net we set $\alpha = 0.5$. We choose to fix the value of $\alpha$ at 0.5 rather than tune it so that we can compare the performance of the equal mixture of the two penalties with using only an $L_1$ or only an $L_2$ penalty. The code used for fitting the nonnegative adaptive lasso is also based on \texttt{glmnet}, were we use 10-fold cross-validation to select both $\lambda$ and $\gamma$. For $\gamma$ we consider the possible values $\{ 0.5, 1, 2\}$, for each of which we fit a path of 100 candidate values for $\lambda$. We then choose the combination of $\gamma$ and $\lambda$ which has the lowest cross-validation error. \par 
For stability selection we use the package \texttt{stabs} 0.6-3 \citep{stabs_package, stabs_article}. We adopt the recommendations of \citet{stabs_article} for choosing the parameters, by specifying $q$ and a desired bound $\textit{PFER}_{\text{max}}$, and then calculating the associated threshold $\pi_{\text{thr}}$. The parameter $q$ should be chosen large enough that in theory all views corresponding to signal can be chosen \citep{stabs_article}. We therefore choose $q = 10$, since we have 10 views corresponding to signal. Note that this means that the procedure has additional information about the true model unavailable to the other meta-learners. We choose a desired bound of $\textit{PFER}_{\text{max}} = 1.5$, which is equivalent to controlling the \textit{per-comparison error rate} (PCER) at $1.5 / 30 = 0.05$ when $V = 30$, or $1.5 / 300 = 0.005$ when $V= 300$. Under the \textit{unimodality assumption} of \citet{Shah2013}, this leads to $\pi_{\text{thr}} = 0.9$ when $V = 30$, and $\pi_{\text{thr}} = 0.57$ when $V = 300$. \par 
The code used to perform nonnegative forward selection is based on \texttt{stepAIC} from \texttt{MASS} 7.3-47 \citep{MASS}. The optimization required for fitting the interpolating predictor is performed using the package \texttt{lsei} 1.2-0 \citep{lsei}. After optimization, coefficients smaller than $\max(10^{-2}/V, 10^{-8})$ are set to zero.

\subsection{Results}

\subsubsection{Effect sizes}

The values of partial $\eta^2$ obtained from the mixed ANOVAs for each of the four outcome measures are given in Table \ref{tab:anova}. Note that we are primarily interested in the extent to which differences between the meta-learners are moderated by the experimental factors of sample size, view size, number of views, and correlation structure. In Table \ref{tab:anova} we therefore show only the interaction terms including the meta-learner factor. \par 
Large or moderate effect sizes can be observed across all four outcome measures for the main effect of the meta-learner, as well as for the interactions with sample size and correlation structure. When accuracy or TPR is used as the outcome, the three-way interaction between meta-learner, sample size and correlation structure also shows a moderate effect size. In Sections \ref{sect:accuracy} through \ref{sect:fdr}, we therefore show the results split by sample size and correlation structure, and use fixed levels of the other experimental factors. In particular, we use $V = 300$ and $m_v = 25$, since this structure is the most similar to our real data examples (Section \ref{sect:realdata}). The results for other combinations of $V$ and $m_v$ can be found in the Appendix. \par 
Note that for the false positive rate only, a moderate effect size can also be observed for the interaction between the choice of meta-learner and the number of views. For the false discovery rate, a borderline moderate effect size can be observed for the three-way interaction between the choice of meta-learner, $V$, and $n$. However, these interactions appear to be dominated by the interpolating predictor. In particular, when $V < n$ the interpolating predictor generally produces sparse models with around 3 nonzero coefficients on average. However, when $V > n$ (i.e. when the view selection problem is high-dimensional) the interpolating predictor produces dense models with around 90 nonzero coefficients, leading to a large increase in FPR and FDR. \par 

\begin{table}[ht]
	\centering
	\caption{Standardized measures of effect size (partial $\eta^2$) for the interactions between the choice of meta-learner and the other experimental factors, for each of the four outcome measures of true positive rate, false positive rate, false discovery rate and classification accuracy. Large effect sizes ($\eta^2 \geq 0.14$) are printed in bold. Moderate effect sizes ($0.06 \leq \eta^2 < 0.14$) are printed in italics. $V$ denotes the number of views, $m_v$ the number of features per view, $n$ the sample size, and cor the correlation structure. \label{tab:anova}}
	\begin{tabular}{lrrrrr}
		\hline 
		& $\eta^2$ (accuracy) & $\eta^2$ (TPR) & $\eta^2$ (FPR) & $\eta^2$ (FDR) \\
		\hline
		meta-learner  & \textbf{.208} & \textbf{.552} & \textbf{.483} & \textbf{.466} \\
		meta-learner*$V$ & .038 & .011 & \textit{.096} & .030  \\
		meta-learner*$m_v$ & .019 & .006 & .011 & .010  \\
		meta-learner*$n$ & \textit{.130} & \textbf{.421} & \textbf{.164} & \textbf{.188}  \\
		meta-learner*cor & \textbf{.236} & \textbf{.386} & \textbf{.243} & \textit{.064}  \\
		meta-learner*$V$*$n$ & .008 & .028 & .045 & \textit{.060} \\
		meta-learner*$v$*cor & .028 & .028 & .040 & .030 \\
		meta-learner*$m_v$*n & .001 & .003 & .003 & .006 \\
		meta-learner*$mv$*cor & .035 & .013 & .009 & .011 \\
		meta-learner*$n$*cor & \textit{.080} & \textit{.094} & .022 & .045 \\
		meta-learner*$V$*$n$*cor & .024 & .010 & .007 & .020 \\
		meta-learner*$m_v$*$n$*cor & .004 & .008 & .004 & .013 \\
		\hline
	\end{tabular}
\end{table}

\newpage
\subsubsection{Test accuracy} \label{sect:accuracy}

\FloatBarrier

\begin{figure}[h!]
	\centering
	\includegraphics{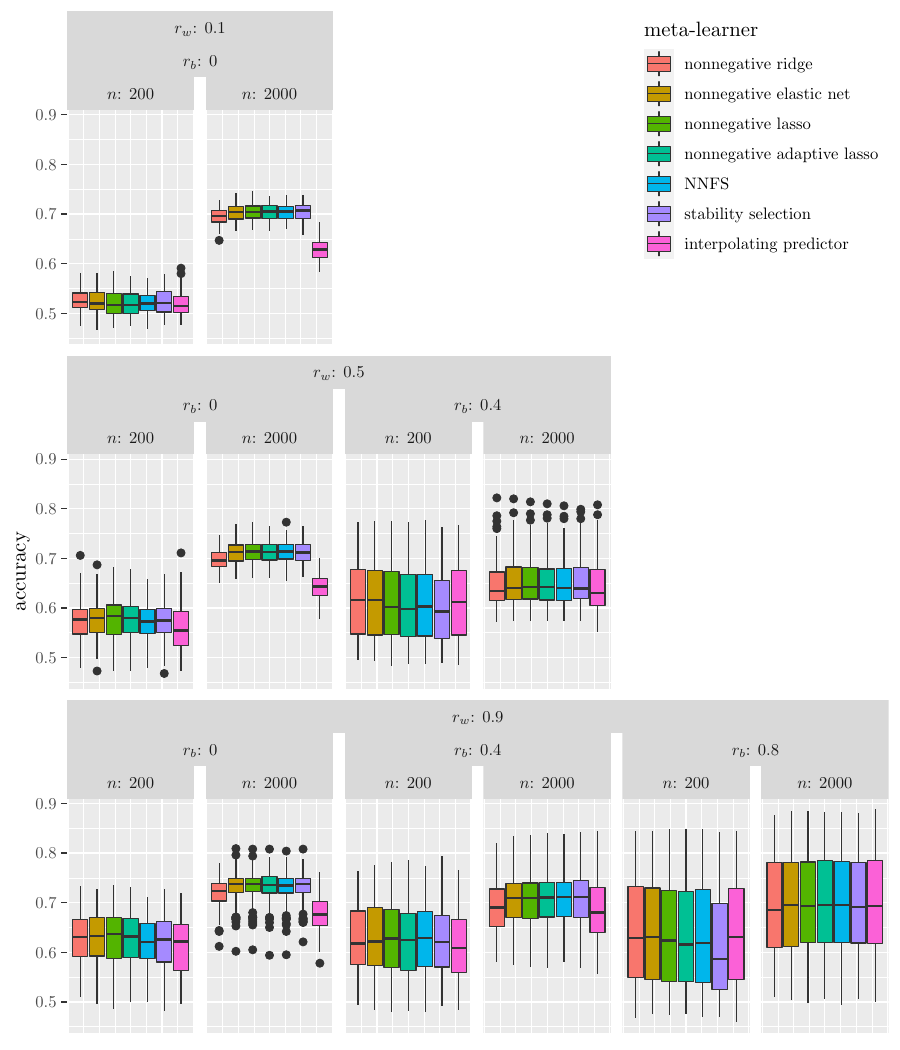}
	\caption{Boxplots of test accuracy for the different meta-learners, with 300 views and 25 features per view. The results are shown for all combinations of the correlation between features within the same view ($\rho_w$), the correlation between features from different views ($\rho_b$), and sample size ($n$). Each plot is based on 100 replications. \label{fig:accuracy}}
\end{figure}

Classification accuracy on the test set for each of the meta-learners can be observed in Figure \ref{fig:accuracy}. Based on these results, the meta-learners can be divided into two groups: On the one hand, the nonnegative lasso, adaptive lasso, elastic net and NNFS generally have very similar classification performance; on the other hand, nonnegative ridge regression, the interpolating predictor and stability selection all perform noticeably worse than the other meta-learners in a subset of the experimental conditions. In particular, ridge regression and the interpolating predictor perform worse when the features in different views are uncorrelated ($\rho_b = 0$), or when the correlation between the features in different views is much lower than the correlation between features in the same view ($\rho_b = 0.4$, $\rho_w = 0.9$). Stability selection performs worse when $n = 200$ and the correlation between features from different views is of a similar magnitude as the correlation between features from the same view (i.e. $\rho_b = 0.4$, $\rho_w = 0.5$ or $\rho_b = 0.8$, $\rho_w = 0.9$). These results appear even more pronounced in the case when $V = 30$, see Figures \ref{fig:acc_v30_mv250} and \ref{fig:acc_v30_mv2500} in the Appendix.   

\FloatBarrier

\newpage
\subsubsection{View selection: True positive rate} \label{sect:tpr}

\FloatBarrier

\begin{figure}[h!]
	\centering
	\includegraphics{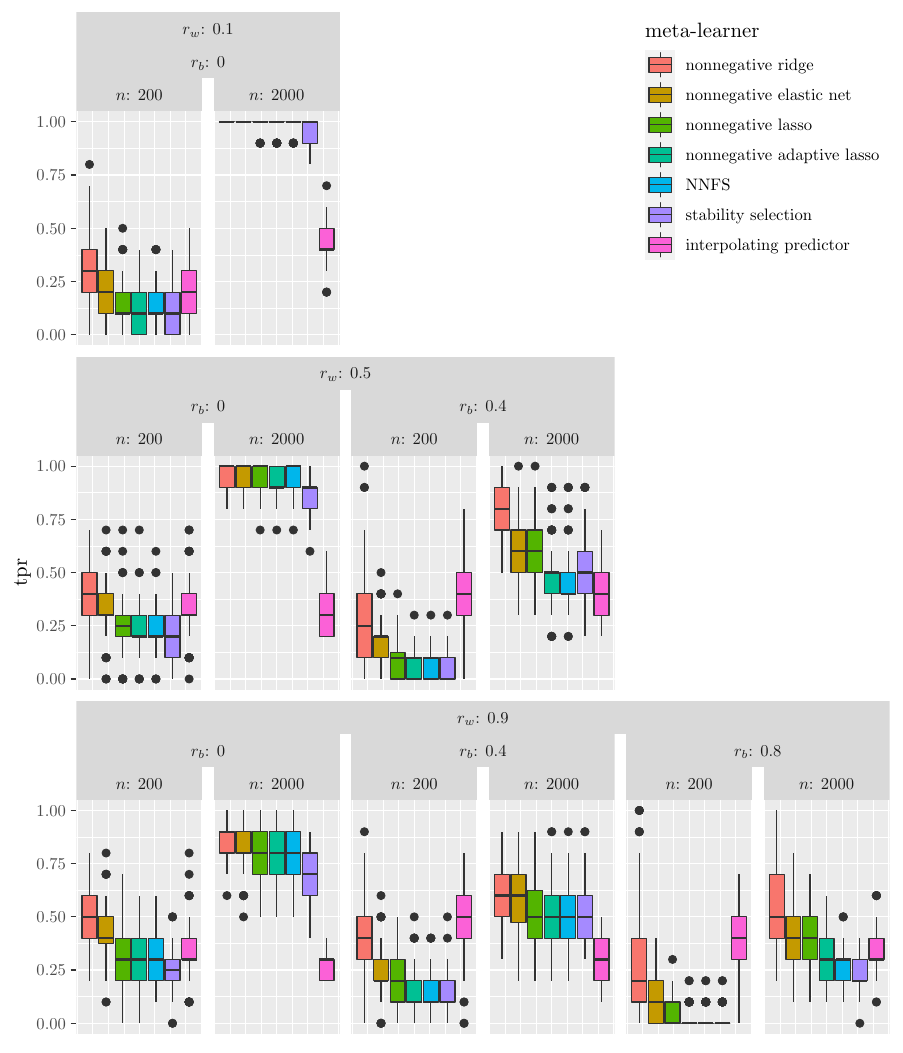}
	\caption{Boxplots of the true positive rate (TPR) for the different meta-learners, with 300 views and 25 features per view. The results are shown for all combinations of the correlation between features within the same view ($\rho_w$), the correlation between features from different views ($\rho_b$), and sample size ($n$). Each plot is based on 100 replications. \label{fig:tpr}}
\end{figure}

The true positive rate in view selection for each of the meta-learners can be observed in Figure \ref{fig:tpr}. Ignoring the interpolating predictor for now, nonnegative ridge regression has the highest TPR, which is unsurprising seeing as it performs feature selection only through its nonnegativity constraints. Nonnegative ridge regression is followed by the elastic net and then lasso. The lasso is followed by the adaptive lasso, NNFS and stability selection, although the order among these three methods changes somewhat for the different conditions. The interpolating predictor shows behavior that is completely different from the other meta-learners. Whereas for the other meta-learners the TPR increases as sample size increases, the TPR of the interpolating predictor actually decreases in some cases. Although it appears to have the highest TPR in some conditions, it can be observed in the next section that it also has the highest FPR in these conditions.

\FloatBarrier

\newpage
\subsubsection{View selection: False positive rate} \label{sect:fdp}

\FloatBarrier

\begin{figure}[h!]
	\centering
	\includegraphics{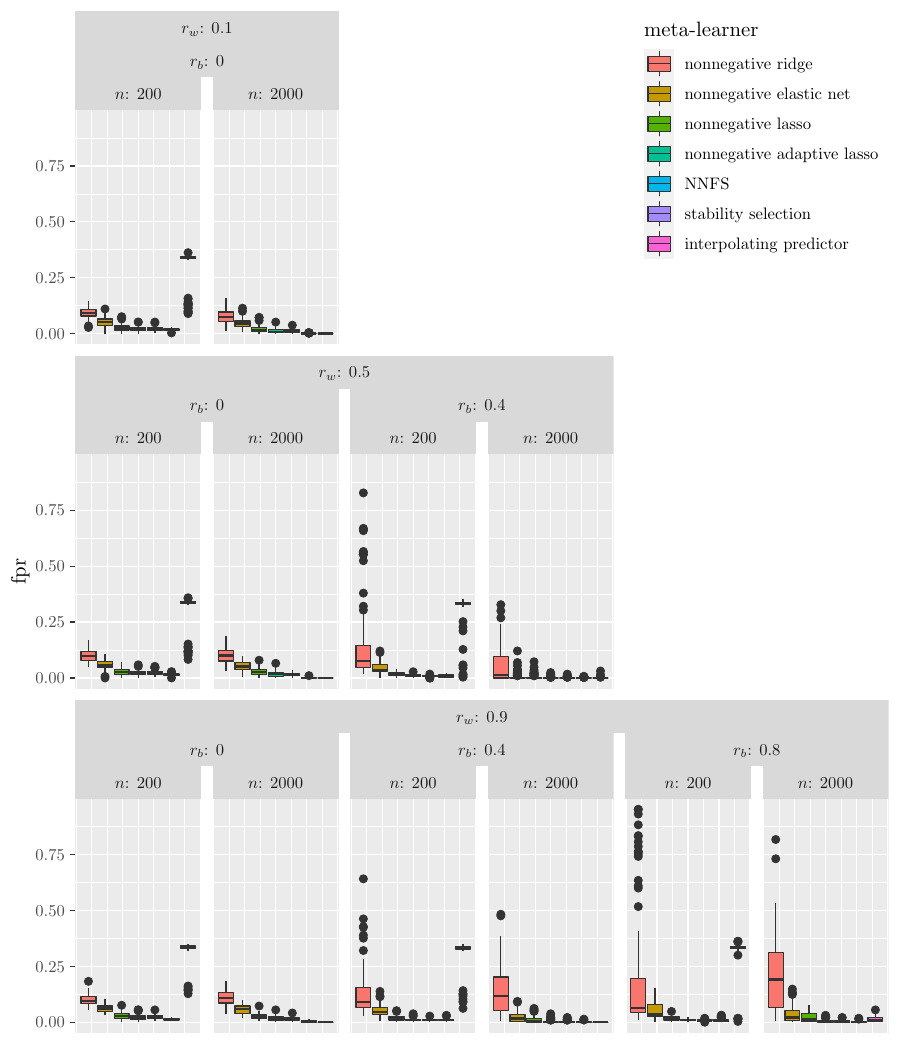}
	\caption{Boxplots of the false positive rate (FPR) for the different meta-learners, with 300 views and 25 features per view. The results are shown for all combinations of the correlation between features within the same view ($\rho_w$), the correlation between features from different views ($\rho_b$), and sample size ($n$). Each plot is based on 100 replications. \label{fig:fpr}}
\end{figure}

The false positive rate in view selection for each of the meta-learners can be observed in Figure \ref{fig:fpr}. Again ignoring the interpolating predictor for now, the ranking of the different meta-learners is similar to their ranking by TPR. Nonnegative ridge regression has the highest FPR, followed by the elastic net, lasso, adaptive lasso and NNFS (which have almost identical performance), and finally stability selection. It is clear from Figure \ref{fig:fpr} that using a meta-learner specifically aimed at view selection can decrease the FPR substantially compared to using only nonnegativity constraints. Again the interpolating predictor shows different behavior. In particular, it has the highest FPR whenever $n = 200$. This appears to be caused by the interpolating predictor producing very dense models whenever the view selection problem is high-dimensional. 

\FloatBarrier

\newpage
\subsubsection{View selection: False discovery rate} \label{sect:fdr}

\FloatBarrier

\begin{figure}[h!]
	\centering
	\includegraphics{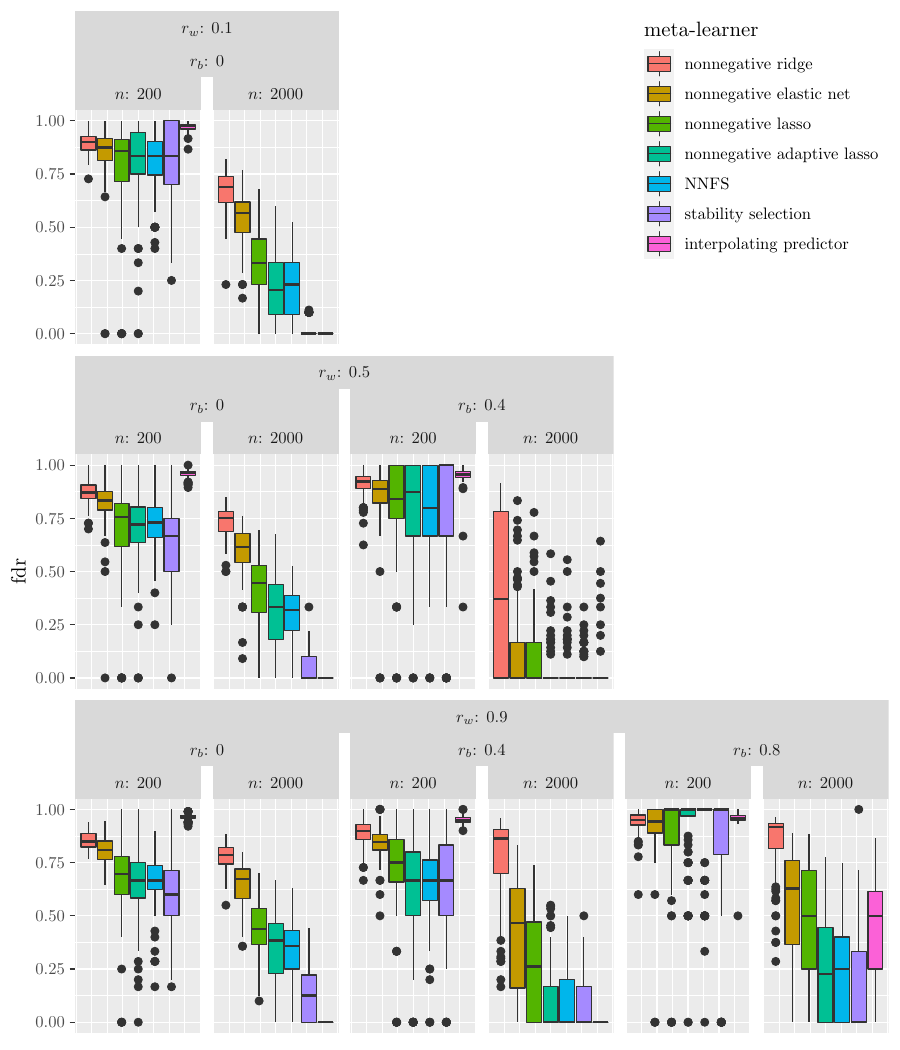}
	\caption{Boxplots of the false discovery rate (FDR) for the different meta-learners, with 300 views and 25 features per view. The results are shown for all combinations of the correlation between features within the same view ($\rho_w$), the correlation between features from different views ($\rho_b$), and sample size ($n$). Each plot is based on 100 replications. \label{fig:fdr}}
\end{figure}

The false discovery rate in view selection for each of the meta-learners can be observed in Figure \ref{fig:fdr}. Note that the FDR is particularly sensitive to variability since its denominator is the number of selected views, which itself is a variable quantity. In particular, when the number of selected views is small, the addition or removal of a single view may cause large increases or decreases in FDR. This happens especially whenever $\rho_b > 0$, as can be observed in Figure \ref{fig:fdr}. The ranking of the different meta-learners is similar to their ranking by TPR and FPR. When $n = 200$, the interpolating predictor has the highest FDR due to its tendency to select very dense models when $n < V$. When $n = 2000$, the interpolating predictor often has a very low FDR, but in these settings it also has considerably lower TPR and test accuracy than the other meta-learners. Of the other meta-learners nonnegative ridge regression has the highest FDR, followed by the elastic net, lasso, adaptive lasso and NNFS, and stability selection.

\FloatBarrier

\subsubsection{Summary of simulation results} \label{sect:summary}

In summary, the nonnegative lasso, adaptive lasso, elastic net and NNFS generally showed comparable classification performance in our simulations, while nonnegative ridge regression, stability selection and the interpolating predictor performed noticeably worse in a subset of the experimental conditions. Among the meta-learners that performed well in terms of accuracy, model sparsity was generally associated with a lower false positive rate in terms of view selection, but also with a lower true positive rate. Nevertheless, there are situations when the sparser meta-learners obtained both a low FPR and a high TPR, particularly when the features from different views were uncorrelated. However, even when the FPR was very low, the FDR was often high, especially in the setting with a sample size of 200.

\FloatBarrier
\section{Gene expression application} \label{sect:realdata}

\subsection{Design}

We apply MVS with the seven different meta-learners to two gene expression data sets, namely the colitis data of \citet{colitis}, and the breast cancer data of \citet{breastcancer}. These data sets were previously used to compare the group lasso with the sparse group lasso \citep{Simon2013}, and to compare the group lasso with StaPLR \citep{StaPLR}. \par 
The colitis data \citep{colitis} consists of 85 colitis cases and 42 healthy controls for which gene expression data was collected using 22,283 probe sets. As in \citep{StaPLR}, we matched this data to the C1 cytogenetic gene sets from MSigDB 6.1 \citep{C1}, and removed any duplicate probes, genes not included in the C1 gene sets, and gene sets which consisted of only a single gene after matching. This led to a multi-view data set consisting of 356 views (gene sets), with an average view size of 33 features (genes). A boxplot of the distribution of the view sizes is included in Appendix \ref{sect:view_sizes}. The total number of features was 11,761. All features were $\text{log}_2$-transformed, then standardized to zero mean and unit variance before applying the MVS procedure. \par 
The breast cancer data \citep{breastcancer} consists of 60 tumor samples labeled according to whether cancer did (28 cases) or did not (32 cases) recur. The data was matched to the C1 gene sets using the same procedure as in the colitis data, leading to a multi-view data set of 354 views, with an average view size of 36 features. A boxplot of the distribution of the view sizes is included in Appendix \ref{sect:view_sizes}. The total number of features was 12,722. The features were already $\text{log}_2$-transformed, but were further standardized to zero mean and unit variance before applying the MVS procedure. \par 
To assess classification performance for each of the data sets, we perform 10-fold cross-validation. We repeat this procedure 10 times and average the results to account for random differences in the cross-validation partitions. The hold-out data in the cross-validation procedure is used only for model evaluation, not for parameter tuning. We again report classification accuracy using a standard threshold of 0.5. Because the colitis data is somewhat unbalanced in terms of class membership, one might additionally be interested in the performance of the methods across multiple possible thresholds. Due to different opinions regarding which metric is most suitable for comparing performance across multiple thresholds \citep{Hand2009,Flach2011,Flach2012}, we report two popular metrics, namely the area under the receiver operating characteristic curve (AUC), and the H measure \citep{Hand2009}. Both metrics can take values in $\lbrack0,1\rbrack$, but the AUC more typically takes values in $\lbrack0.5,1\rbrack$, with a value of 0.5 denoting a noninformative classifier. For the H measure, a noninformative classifier is associated with a value of zero. \par    
In terms of view selection, each of the $10 \times 10$ fitted models is associated with a set of selected views. However, quantities like TPR, FPR and FDR cannot be computed since the true status of the views is unknown. We therefore report the number of selected views, since this allows assessment of model sparsity. In addition, we report a measure of the stability of the set of selected views. In particular, we use the feature selection stability measure of \citet{Nogueira2018}.  This stability measure, $\hat{\Phi}$, has an upper bound of 1 and a lower bound which is asymptotically zero but depends on the number of fitted models (in our case it is approximately -0.01). Higher values indicate increased stability, and  $\hat{\Phi}$ attains its maximum of 1 if-and-only-if the set of selected views is the same for each fitted model \citep{Nogueira2018}. A particularly desirable property of this stability measure is that it is corrected for chance in the sense that its expected value is zero if a selection algorithm would select sets of views of a certain size randomly rather than systematically \citep{Nogueira2018}. The measure can additionally be considered a special case of the Fleiss' Kappa \citep{Fleiss1971} measure of inter-rater agreement, where each of the fitted models is a ``rater" classifying the views into whether or not they are relevant \citep{Nogueira2018}. Rules of thumb for interpreting the strength of agreement associated with a certain value of Fleiss' Kappa have been formulated by \citet{Fleiss_interpretation}: .00-.20=slight, 0.21-0.40=fair, 0.41-0.60=moderate, 0.61-0.80=substantial, 0.81-1.0=almost perfect. However, these rules of thumb are largely arbitrary \citep{Fleiss_interpretation}, and we will focus on relative comparisons. \par 

\subsection{Software}

We use the same software as described in Section \ref{sect:software}. All cross-validation loops used for parameter tuning are nested within the outer loop used for evaluating classification performance. We again use the recommendations of \citet{stabs_article} for choosing the parameters, by specifying $q$ and a desired bound $\textit{PFER}_{\text{max}}$, and then calculating the associated threshold $\pi_{\text{thr}}$. We again specify a desired bound of 1.5. The parameter $q$ should be large enough so that all views corresponding to signal can be selected \citep{stabs_article}, but in real data the number of views corresponding to signal is unknown. However, domain knowledge or results from previous experiments can be used to obtain an estimate of the number of views corresponding to signal, and thus assist in choosing the value of $q$. The colitis and breast cancer data sets have previously been analyzed using StaPLR with the nonnegative lasso as a meta-learner \citep{StaPLR}. We set the value of $q$ to the maximum number of views selected by the StaPLR method in the colitis and breast cancer data sets as reported in \citet{StaPLR}. This leads to $q = 15$, $\pi_{\text{thr}} = 0.62$ for the colitis data, and $q = 11$, $\pi_{\text{thr}} = 0.57$ for the breast cancer data. Note that this means the stability selection procedure has access to information from a previous experiment that the other meta-learners do not have access to. The AUC is calculated using the package \texttt{AUC} 0.3.0 \citep{AUC}, and the H measure using the package \texttt{hmeasure} 1.0-2 \citep{hmeasure}. The stability measure $\hat{\Phi}$ was calculated using a custom script, which is included in the supplementary materials.

\subsection{Results}

The results of applying MVS with the seven different meta-learners to the colitis data can be observed in Table \ref{tab:colitis}. In terms of raw test accuracy the nonnegative lasso is the best performing meta-learner, followed by the nonnegative elastic net and the nonnegative adaptive lasso. In terms of AUC and H, the best performing meta-learners are the elastic net and ridge regression. However, the elastic net selects on average almost 4 times as many views as the lasso, and nonnegative ridge regression selects on average almost 13 times as many views, and both have lower raw test accuracy than the lasso. \par 
Since feature selection often comes at a cost in terms of stability \citep{Xu2012}, it is to be expected that view selection stability ($\hat{\Phi}$) is higher for meta-learners that select more views. The results of two meta-learners do not align with this pattern, namely those for the interpolating predictor and NNFS. The interpolating predictor is very dense but has lower stability than several sparse models, while NNFS is less sparse than stability selection and also less stable. Note that we mean sparsity in terms of the number of selected views, but this corresponds to sparsity in terms of the number of selected features (see Appendix \ref{sect:view_sizes}, Table \ref{tab:features}). \par 
The results for the breast cancer data can be observed in Table \ref{tab:breast_cancer}. The interpolating predictor and the lasso are the best performing meta-learners in terms of all three classification measures, with the interpolating predictor having higher test accuracy and H, and the lasso having higher AUC. However, the interpolating predictor selects over 80 times as many views as the lasso, and is less stable. Again, the interpolating predictor and NNFS do not align with the pattern that less sparsity is associated with higher stability.     

\FloatBarrier

\begin{table}[ht]
	\centering
		\caption{Results of applying MVS with different meta-learners to the colitis data. ANSV denotes the average number of selected views. H denotes the H measure \citep{Hand2009}. In computing the H measure we assume that the misclassification cost is the same for each class. $\hat{\Phi}$ denotes the feature selection stability measure of \citet{Nogueira2018}. For accuracy, AUC and H we show the mean and standard deviation across the 10 replications. For the number of selected views we show the mean and standard deviation across the $10 \times 10$ different fitted models. The total number of views for this data set is 356, and 67\% of observations belong to the majority class.} \label{tab:colitis}
	\begin{tabular}{rrrrrrr}
		\hline
		Meta-learner & Accuracy & AUC & H & ANSV & $\hat{\Phi}$ \\
		\hline
		lasso & $\textbf{.963} \pm .010$  & $.985 \pm .009$ & $.892 \pm .027$ & $10.7 \pm 1.9$ & .539 \\
		elastic net & $.956 \pm .008$  & $.989 \pm .006$ & $\textbf{.896} \pm .017$ & $40.0 \pm 4.6$ & .606 \\
		adaptive lasso & $.955 \pm .015$ &  $.981 \pm .010$ & $.883 \pm .028$ & $7.4 \pm 1.6$ & .488 \\
		ridge & $.948 \pm .008$ &  $\textbf{.992} \pm .003$  & $.894 \pm .014$ & $132.7 \pm 18.1$ & \textbf{.641} \\
		interpolating predictor & .$940 \pm .006$ &  $.988 \pm .001$ & $.863 \pm .010$ & $242.0 \pm 1.6$ & .463 \\
		NNFS & $.928 \pm .019$ &  $.943 \pm .026$ & $.788 \pm .063$ & $2.6 \pm 0.5$ & .198 \\
		stability selection & $.923 \pm .013$  & $.951 \pm .011$ & $.788 \pm .040$ & $2.0 \pm 0.9$ & .375 \\
		\hline
	\end{tabular}
\end{table}

\begin{table}[ht]
	\centering
	\caption{Results of applying MVS with different meta-learners to the breast cancer data. ANSV denotes the average number of selected views. H denotes the H measure \citep{Hand2009}. In computing the H measure we assume that the misclassification cost is the same for each class. $\hat{\Phi}$ denotes the feature selection stability measure of \citet{Nogueira2018}. For accuracy, AUC and H we show the mean and standard deviation across the 10 replications. For the number of selected views we show the mean and standard deviation across the $10 \times 10$ different fitted models. The total number of views for this data set is 354, and 53\% of observations belong to the majority class.} \label{tab:breast_cancer}
	\begin{tabular}{rrrrrrr}
		\hline
		Meta-learner & Accuracy & AUC & H & ANSV & $\hat{\Phi}$ \\
		\hline
		lasso & $.660 \pm .031$ &  $\textbf{.681} \pm .024$ & $.236 \pm .047$ & $3.6 \pm 1.8$ & .334 \\
		elastic net & $.647 \pm .037$ & $.665 \pm .026$ & $.201 \pm .033$ & $13.1 \pm 2.8$ & .396 \\
		adaptive lasso & $.652 \pm .034$ & $.675 \pm .034$ & $.222 \pm .058$ & $2.6 \pm 1.2$ & .313 \\
		ridge & $.653 \pm .031$ & $.672 \pm .021$ & $.207 \pm .024$ & $52.1 \pm 15.9$ & \textbf{.476} \\
		interpolating predictor & $\textbf{.682} \pm .038$  & $.676 \pm .023$ & $\textbf{.239} \pm .060$ & $298.6 \pm 1.6$ & .118 \\
		NNFS & $.633 \pm .024$ & $.644 \pm .045$ & $.162 \pm .058$ & $3.1 \pm 1.0$ & .257 \\
		stability selection & $.540 \pm .045$ & $.542 \pm .045$ & $.073 \pm .042$ & $1.5 \pm 1.1$ & .108 \\ 
		\hline
	\end{tabular}
\end{table}

\FloatBarrier

\section{Discussion}

In this article we investigated how different view-selecting meta-learners affect the performance of multi-view stacking. In our simulations, the interpolating predictor often performed worse than the other meta-learners on at least one outcome measure. For example, when the sample size was larger than the number of views, the interpolating predictor often had the lowest TPR in view selection, as well as the lowest test accuracy, particularly when there was no correlation between the different views. When the sample size was smaller than the number of views, the interpolating predictor had a FPR in view selection that was considerably higher than that of all other meta-learners. In terms of accuracy it performed very well in the breast cancer data, but less so in the colitis data. However, in both cases it produced very dense models, which additionally had low view selection stability. The fact that its behavior varied considerably across our experimental conditions, combined with its tendency to select very dense models when the meta-learning problem is high-dimensional, suggests that the interpolating predictor should not be used when view selection is among the goals of the study under consideration. However, it may have some use when its interpretation as a weighted mean of the view-specific models is of particular importance. \par
Excluding the interpolating predictor, nonnegative ridge regression produced the least sparse models. This is not surprising considering it performs view selection only through its nonnegativity constraints. Its high FPR in view selection appeared to negatively influence its test accuracy, as there was generally at least one sparser model with better accuracy in both our simulations and real data examples. Although nonnegative ridge regression shows that the nonnegativy constrains alone already cause many coefficients to be set to zero, if one assumes the true underlying model to be sparse, one should probably choose one of the meta-learners specifically aimed at view selection. \par 
The nonnegative elastic net, with its additional $L_1$ penalty compared with ridge regression, is one such method. In our simulations it produced sparser models than nonnegative ridge regression, usually with better or comparable accuracy. These sparser models were associated with a reduction in FPR and FDR, but in some setting also with a reduction in TPR, particularly when there are correlations between the views. However, we fixed the mixing parameter $\alpha$ at 0.5 to observe a specific setting in between ridge regression and the lasso. In practice, one can tune $\alpha$, for example through cross-validation. This may allow the elastic net to better adapt to different correlation structures. In the colitis data, the elastic net performed better than nonnegative ridge regression in terms of test accuracy, whereas in the breast cancer data it performed slightly worse. However, in both cases it produced much sparser models, demonstrating its use in view selection. \par 
The nonnegative lasso, utilizing only an $L_1$ penalty, produced even sparser models than the elastic net. Interestingly, in our simulations this increased sparsity did not appear to have a substantial negative effect on accuracy, although some minor reductions were observed in some low sample size cases. In the colitis data it performed best in terms of raw test accuracy, second in terms of $H$ measure, and third in terms of AUC. In the breast cancer data it performed second-best in accuracy and $H$ measure, and best in terms of AUC. Notably, it selected on average only 3.7 views out of 354, whereas the only better performing meta-learner in terms of accuracy, the interpolating predictor, selected on average 298.6 views. Our results indicate that using the nonnegative lasso as a meta-learner can substantially reduce the number of views while still providing accurate prediction models. \par 
Our implementation of the nonnegative adaptive lasso produced slightly sparser models than the regular nonnegative lasso. This did not appear to substantially reduce classification accuracy in our simulations, although there were some minor reductions in some low sample size cases. In both gene expression data sets the adaptive lasso performed worse on average than the lasso in all three classification metrics, but the observed differences were small. The main difference between these two meta-learners appears to be that the regular lasso slightly favors classification performance, whereas the adaptive lasso slightly favors sparsity. Note that the adaptive lasso is a flexible method, and one can change the way in which its weights are initialized, which will likely affect performance. Additionally, one could consider a larger set of possible values for the tuning parameter $\gamma$. However, this flexibility also means the method is less straightforward to use than the regular lasso. \par 
The NNFS algorithm performed surprisingly well in our simulations given its simple and greedy nature, showing performance very similar to that of the adaptive lasso. However, in both gene expression data sets it was among the two worst performing methods, both in terms of accuracy and view selection stability. If one additionally considers that NNFS does not scale well with larger problems there is generally no reason to choose this algorithm over the nonnegative (adaptive) lasso. \par 
Excluding the interpolating predictor, stability selection produced the sparsest models in our simulations. However, this led to a reduction in accuracy whenever the correlation within features from the same view was of a similar magnitude as the correlations between features from different views. In both gene expression data sets stability selection also produced the sparsest models, but it also had the worst classification accuracy of all meta-learners. In applying stability selection, one has to specify several parameters. We calculated the values of these parameters in part by specifying a desired bound on the PFER (in our case 1.5). This kind of error control is much less strict than the typical family-wise error rate (FWER) or FDR control one would apply when doing statistical inference. In fact, one can observe in Figures \ref{fig:fpr} and \ref{fig:fdr} that although stability selection has a low FPR, for a sample size of 200 its FDR is still much higher than one would typically consider acceptable when doing inference (common FDR control levels are 0.05 or 0.1). Additionally, we gave the meta-learner information about the number of views containing signal in the data (parameter $q$), which the other meta-learners did not have access to. It is also worth noting that the sets of views selected by stability selection in both gene expression data sets had low view selection stability. Ideally, selecting views based on their stability would lead to a set of selected views that is itself highly stable, but evidently this is not the case. It follows then that stability selection may produce a set of selected views which is neither particularly useful for prediction, nor for inference. One could add additional assumptions \citep{Shah2013}, which may increase predictive performance, but may also increase FDR. Or one could opt for stricter error control, but this would likely reduce classification performance even further. This implies that performing view selection for both the aims of prediction and inference using a single procedure may produce poor results, since the resulting set of selected views may not be suitable for either purpose. \par 
In this study we only considered different meta-learners within the MVS framework. Of course, many other algorithms for training classifiers exist. Some of those classifiers may be expected to perform better in terms of classification performance than the classifiers presented here, but not many have the embedded view selection properties of MVS-based methods. For example, a random forest would probably perform very well in terms of classification, but the resulting classifier is hard to interpret and does not automatically select the most important views for prediction. One non-MVS method which does automatically select views is the group lasso \citep{grouplasso}, but we did not include it here as an extensive comparison between StaPLR/MVS and the group lasso has already been performed elsewhere \citep{StaPLR}. \par
Any simulation study is limited by its choice of experimental factors. In particular, in our simulations we assumed that all features corresponding to signal have the same regression weight, and that all views contain an equal number of features. The correlation structures we used are likely simpler than those encountered in real data sets. Additionally, we defined the view selection problem in such a way that we want to select any view which contains at least some (in our simulations at least 50\%) features truly related to the outcome. In practice, the amount of signal present in a view may be lower, leading to considerations of exactly how much signal should be present in a view in order for the researcher to be considered worth selecting. Additionally, we only considered settings where views are mutually exclusive, but in practice views may overlap \citep{Yuan2011,Park2015}, meaning that a single feature may correspond to multiple views. In general, the MVS algorithm can handle overlapping views by simply `copying' a feature for each additional view in which it occurs. However, an exploration of the implications of overlapping views for view selection, both in MVS and in general, would make an interesting topic for future research. We also did not include the possibility of missing data. In multi-view data, it is quite likely that if missing data occurs, all features within a view will be simultaneously missing. Future work may focus on developing optimal strategies for handling missing data in the multi-view context. \par 
In this study, we evaluated the performance of the different meta-learners across a variety of settings, including high-dimensional and highly correlated settings. Most of these settings were not easy problems, as evident by the absolute accuracy values obtained by the meta-learners. Additionally we considered two real data examples, one considerably harder than the other. Across all our experiments, the relative performance of the nonnegative lasso, nonnegative adaptive lasso and nonnegative elastic net remained remarkably stable. Our results show that MVS can be used with one of these meta-learners to obtain models which are substantially sparser at the view level than those obtained with other meta-learners, without incurring a major penalty in classification accuracy.
The nonnegative elastic net is particularly suitable if it is important to the research that, out of a set of correlated features, more than one should be selected. If this is not of particular importance, the nonnegative lasso and nonnegative adaptive lasso can provide even sparser models. \par

\section{Declarations}

\subsection{Funding}


\textit{This information is blinded for review.}

\subsection{Conflicts of interest}

The authors declare no conflicts of interest.

\subsection{Availability of data and material}

The following data sets were used:

\begin{itemize}
	\item Breast cancer data \citep{breastcancer} - NCBI GEO DataSet Record GDS806
	\item Colitis data \citep{colitis} - NCBI GEO DataSet Record GDS1615
	\item C1 gene sets reference file v6.1 \citep{C1} - Available from: \\ http://software.broadinstitute.org/gsea/downloads\_archive.jsp
\end{itemize}

\subsection{Code availability}

The used code is included in the supplementary materials.

\subsection{Authors' contributions}


\textit{This information is blinded for review.}

\section{MSC Codes}

62 - Statistics, 68 - Computer Science

\newpage
\appendix
\FloatBarrier

\section{Additional information on colitis and breast cancer data} \label{sect:view_sizes}

\begin{figure}[h!]
	\centering
	\includegraphics{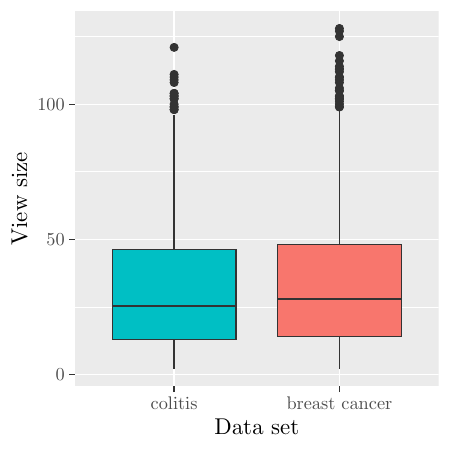}
	\caption{Boxplots of the number of features per view for the breast cancer and colitis data sets.}
\end{figure}

\begin{table}[ht]
	\centering
	\caption{Comparing the view and feature dimension resulting from applying MVS with different meta-learners to the breast cancer and colitis data. ANSV denotes the average number of selected views, ANSF denotes the average number of selected features. We show the mean and standard deviation across the $10 \times 10$ different fitted models.} \label{tab:features}
	\begin{tabular}{l|rr|rr}
		\hline
		& \multicolumn{2}{c|}{Colitis} & \multicolumn{2}{c}{Breast cancer}  \\
		\hline
		Metalearner & ANSV & ANSF & ANSV  & ANSF \\
		\hline
		lasso & $10.7 \pm 1.9$ & $678.4 \pm 119.6$ & $3.6 \pm 1.8$ & $ 99.3 \pm 65.0$ \\
		elastic net & $40.0 \pm 4.6$ & $1978.2 \pm 282.8$ & $13.1 \pm 2.8$ & $388.6 \pm 115.5$ \\
		adaptive lasso & $7.4 \pm 1.6$ & $482.6 \pm 90.9$ & $2.6 \pm 1.2$ & $71.4 \pm 42.0$ \\
		ridge & $132.7 \pm 18.1$ & $5349.6 \pm 749.6$ & $52.1 \pm 15.9$ & $1643.8 \pm 489.3$ \\
		interpolating predictor & $242.0 \pm 1.6$ & $9418.2 \pm 114.6$ & $298.6 \pm 1.6$ & $9778.9 \pm 172.1$ \\
		NNFS & $2.6 \pm 0.5$ & $180.1 \pm 49.7$ & $3.1 \pm 1.0$ & $99.3 \pm 56.6$ \\
		stability selection & $2.0 \pm 0.9$ & $161.7 \pm 76.9$ & $1.5 \pm 1.1$ & $46.7 \pm 49.6$ \\
		\hline
	\end{tabular}
\end{table}

\FloatBarrier
\newpage

\section{Simulation results for $V = 300$ and $m_v = 250$} 

\begin{figure}[h!]
	\centering
	\includegraphics{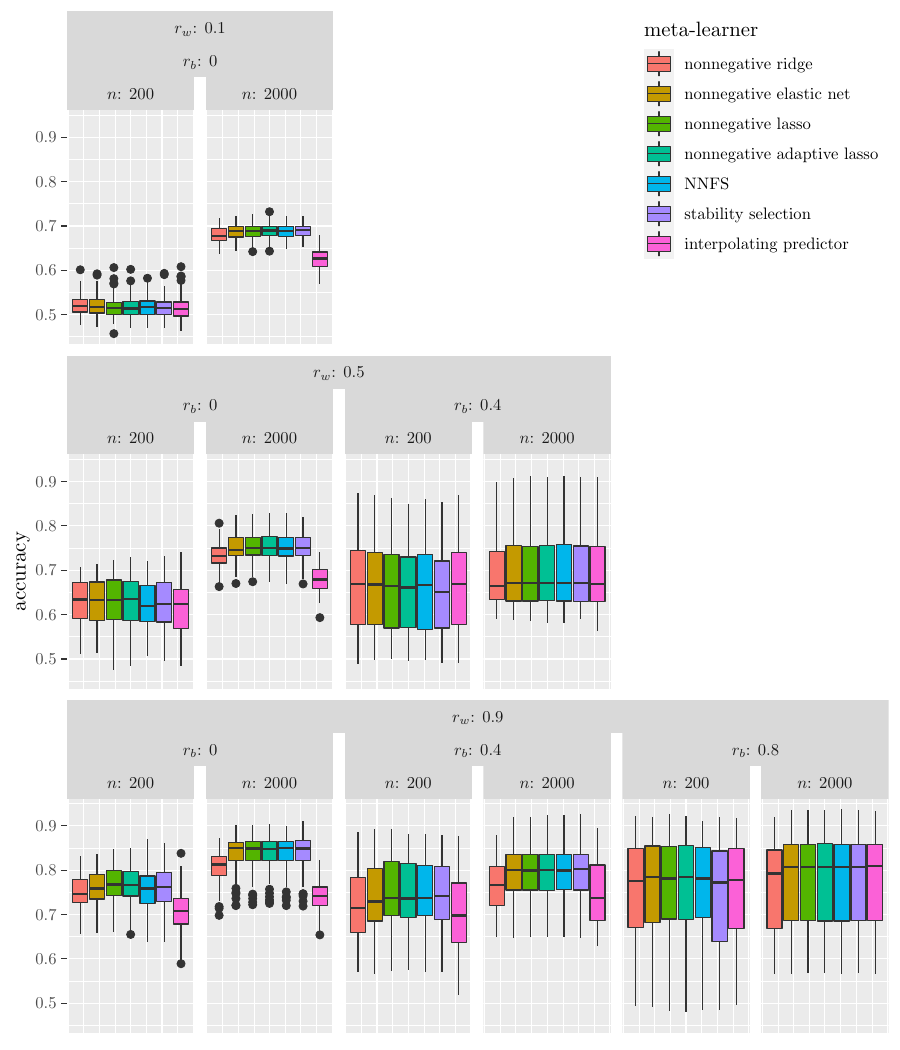}
	\caption{Boxplots of test accuracy for the different meta-learners, with 300 views and 250 features per view. The results are shown for all combinations of the correlation between features within the same view ($\rho_w$), the correlation between features from different views ($\rho_b$), and sample size ($n$). Each plot is based on 100 replications.}
\end{figure}

\begin{figure}[h!]
	\centering
	\includegraphics{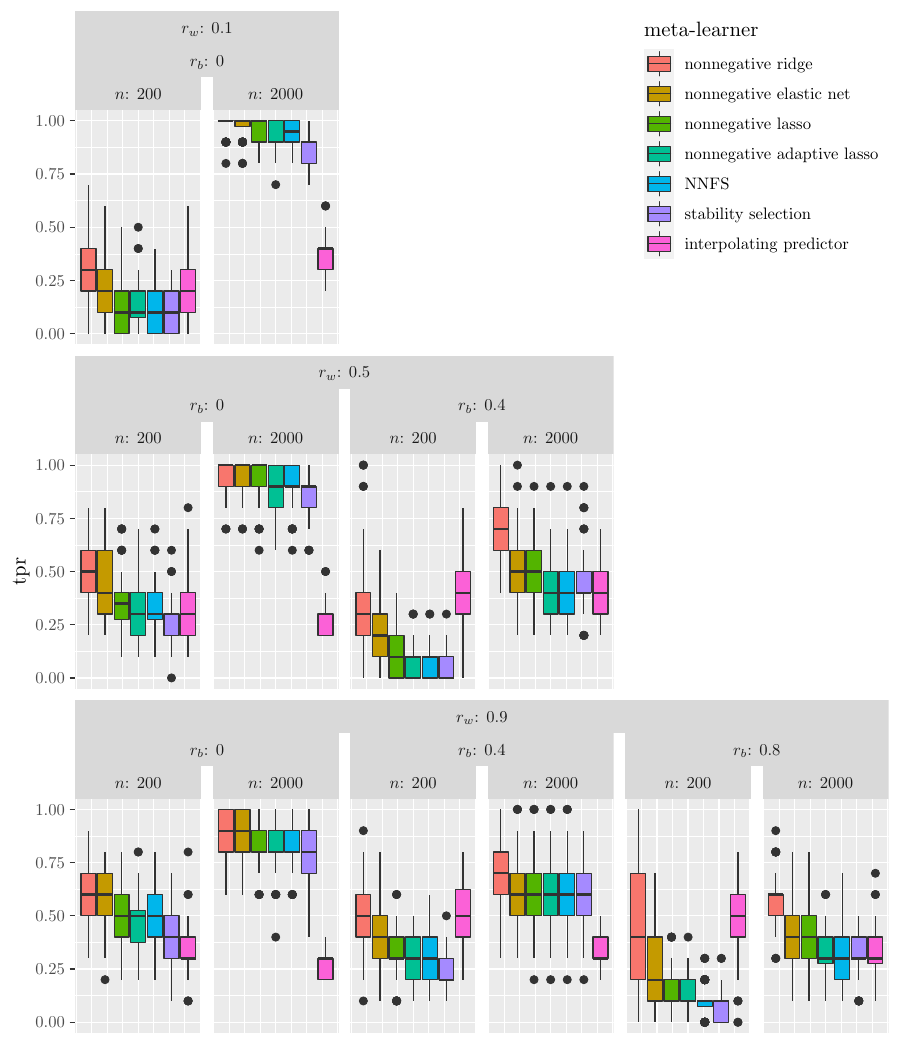}
	\caption{Boxplots of the true positive rate (TPR) for the different meta-learners, with 300 views and 250 features per view. The results are shown for all combinations of the correlation between features within the same view ($\rho_w$), the correlation between features from different views ($\rho_b$), and sample size ($n$). Each plot is based on 100 replications.}
\end{figure}

\begin{figure}[h!]
	\centering
	\includegraphics{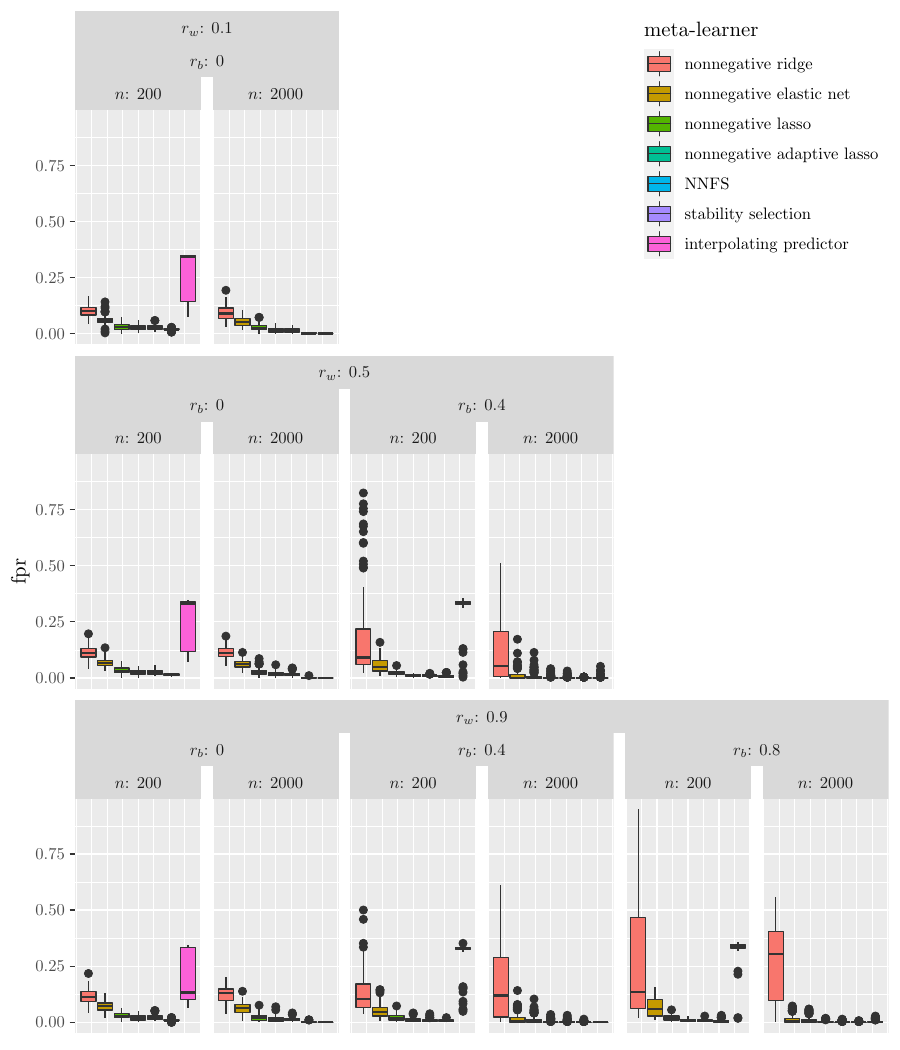}
	\caption{Boxplots of the false positive rate (FPR) for the different meta-learners, with 300 views and 250 features per view. The results are shown for all combinations of the correlation between features within the same view ($\rho_w$), the correlation between features from different views ($\rho_b$), and sample size ($n$). Each plot is based on 100 replications.}
\end{figure}

\begin{figure}[h!]
	\centering
	\includegraphics{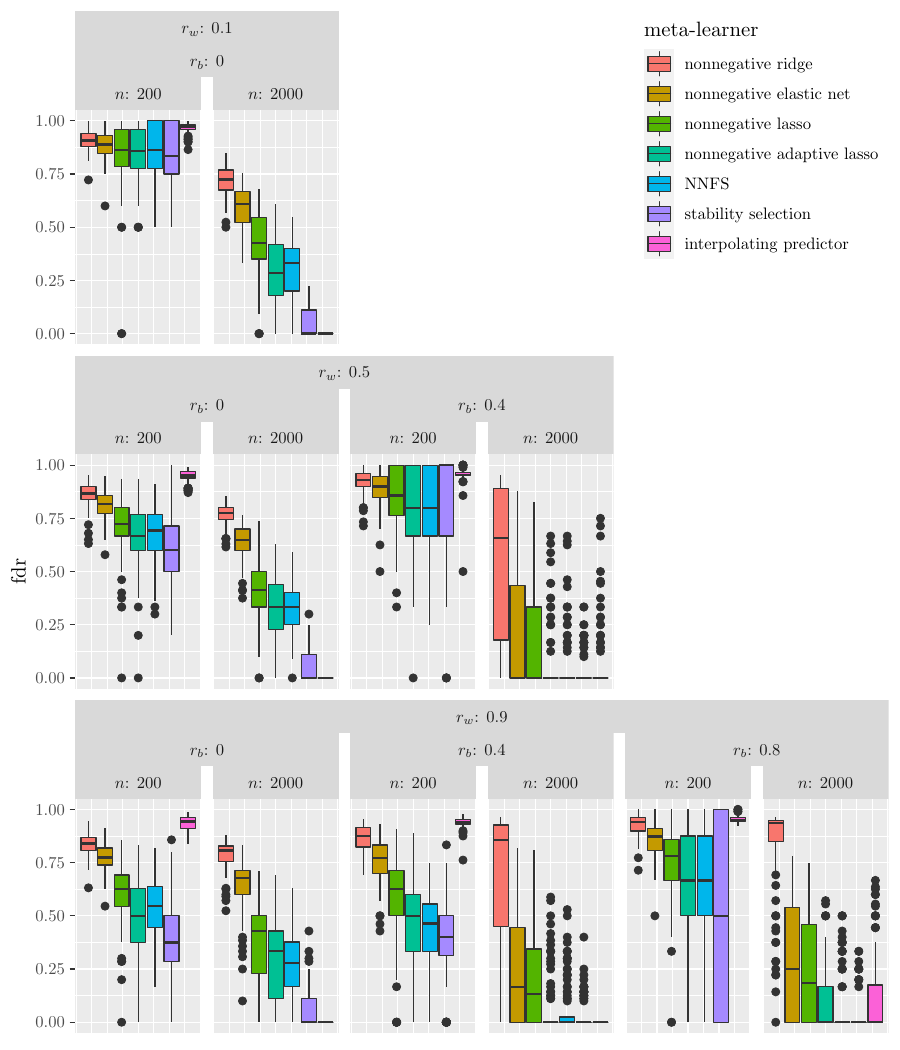}
	\caption{Boxplots of the false discovery rate (FDR) for the different meta-learners, with 300 views and 250 features per view. The results are shown for all combinations of the correlation between features within the same view ($\rho_w$), the correlation between features from different views ($\rho_b$), and sample size ($n$). Each plot is based on 100 replications.}
\end{figure}

\FloatBarrier

\section{Simulation results for $V = 30$ and $m_v = 250$}

\begin{figure}[h!]
	\centering
	\includegraphics{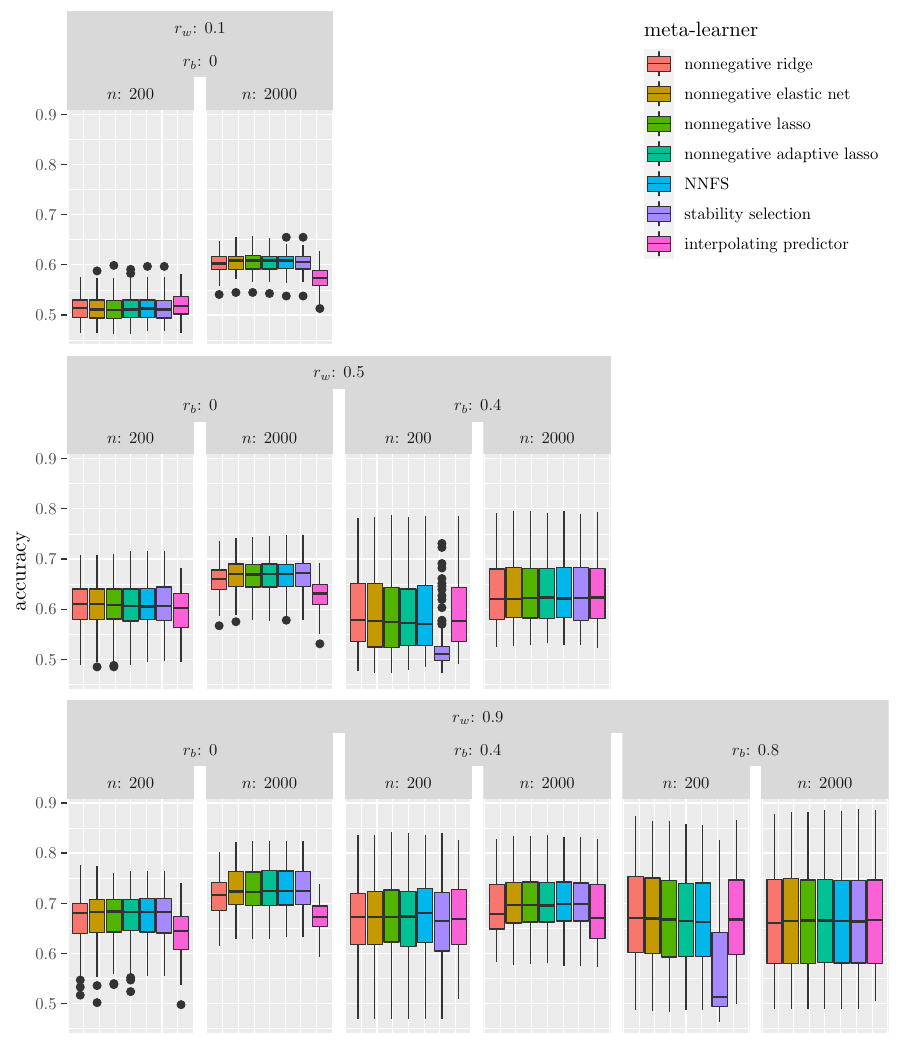}
	\caption{Boxplots of test accuracy for the different meta-learners, with 30 views and 250 features per view. The results are shown for all combinations of the correlation between features within the same view ($\rho_w$), the correlation between features from different views ($\rho_b$), and sample size ($n$). Each plot is based on 100 replications.\label{fig:acc_v30_mv250}}
\end{figure}

\begin{figure}[h!]
	\centering
	\includegraphics{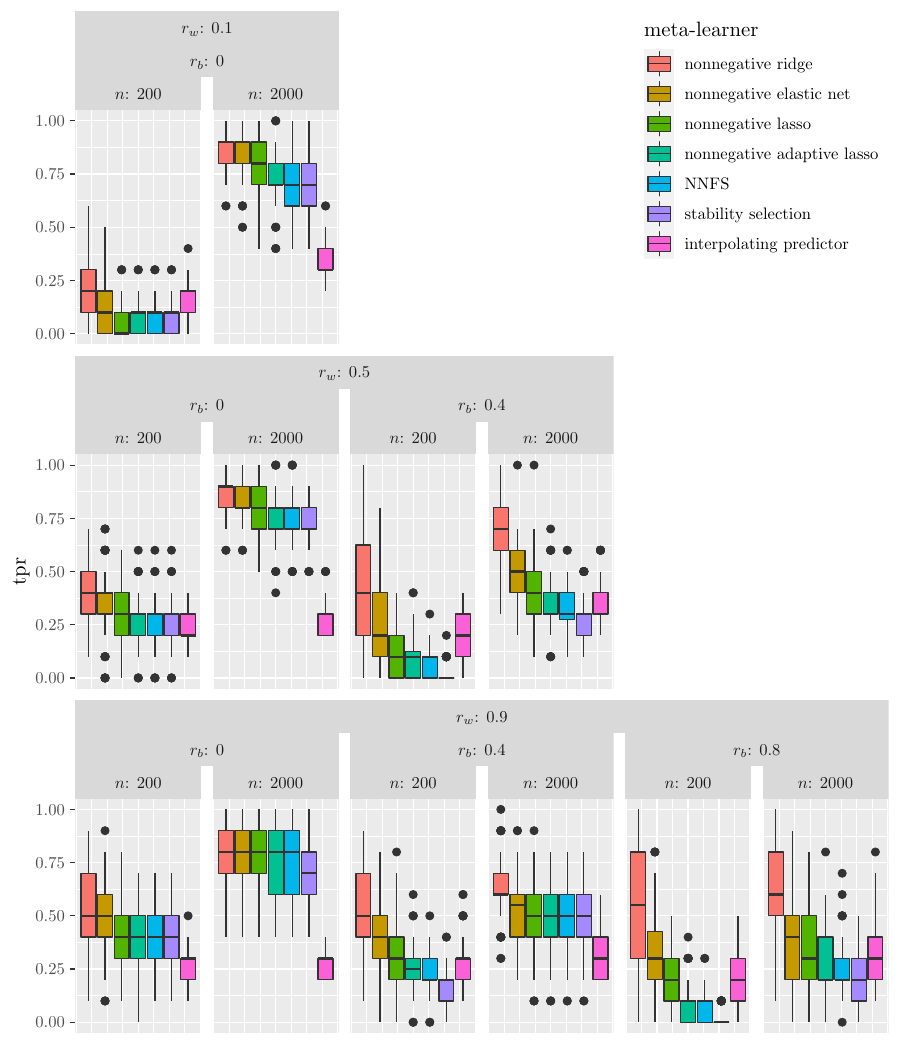}
	\caption{Boxplots of the true positive rate (TPR) for the different meta-learners, with 30 views and 250 features per view. The results are shown for all combinations of the correlation between features within the same view ($\rho_w$), the correlation between features from different views ($\rho_b$), and sample size ($n$).Each plot is based on 100 replications.}
\end{figure}

\begin{figure}[h!]
	\centering
	\includegraphics{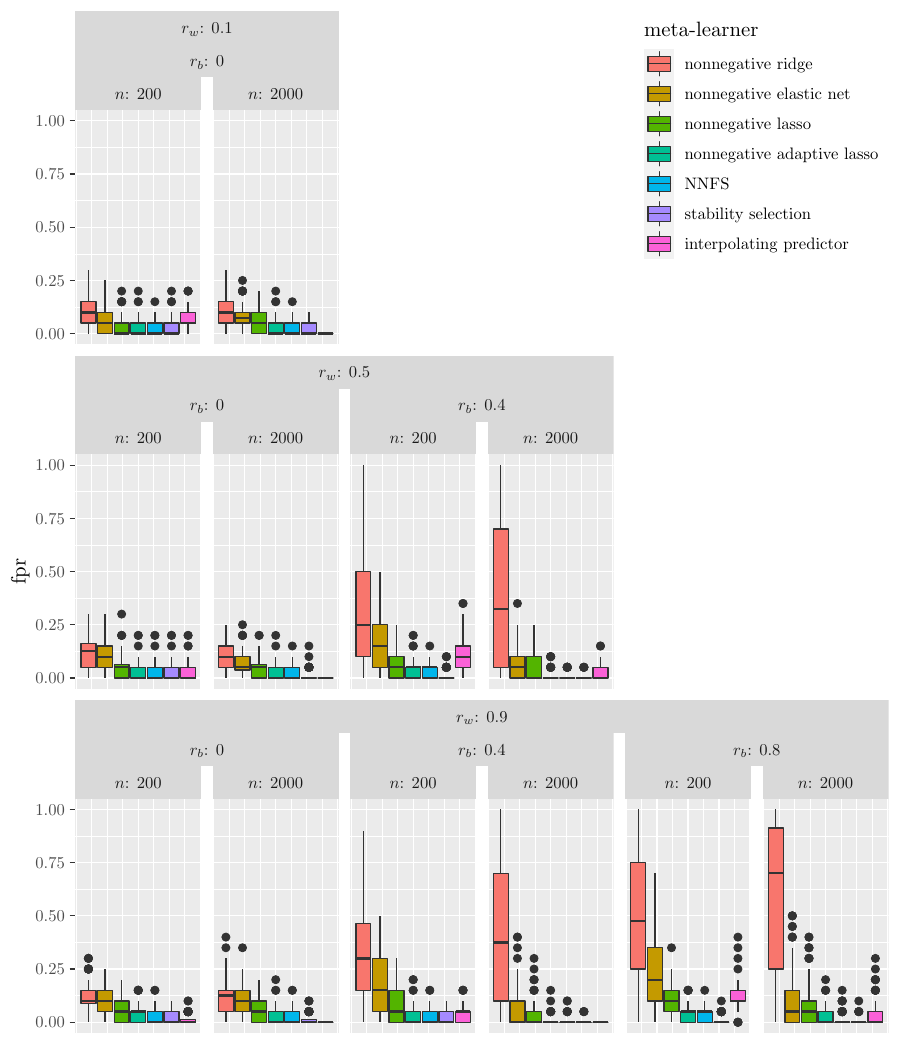}
	\caption{Boxplots of the false positive rate (FPR) for the different meta-learners, with 30 views and 250 features per view. The results are shown for all combinations of the correlation between features within the same view ($\rho_w$), the correlation between features from different views ($\rho_b$), and sample size ($n$).Each plot is based on 100 replications.}
\end{figure}

\begin{figure}[h!]
	\centering
	\includegraphics{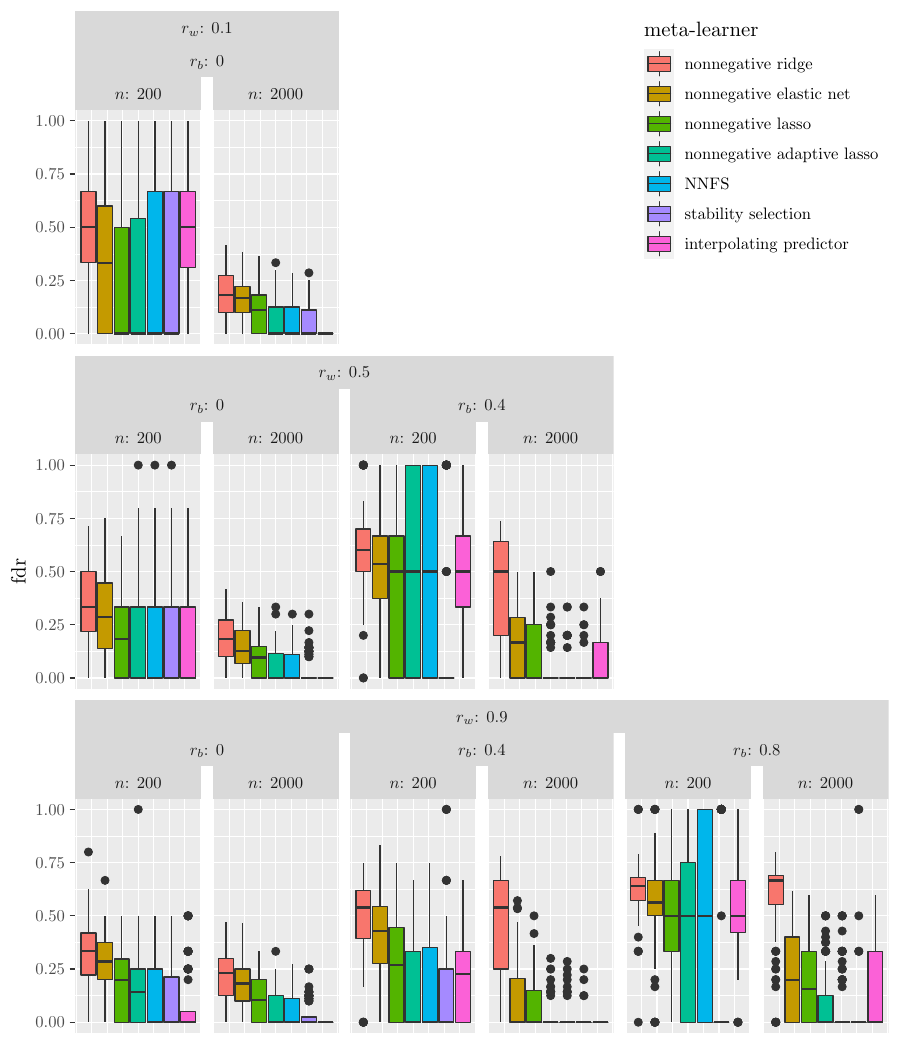}
	\caption{Boxplots of the false discovery rate (FDR) for the different meta-learners, with 30 views and 250 features per view. The results are shown for all combinations of the correlation between features within the same view ($\rho_w$), the correlation between features from different views ($\rho_b$), and sample size ($n$).Each plot is based on 100 replications.}
\end{figure}

\FloatBarrier

\section{Simulation results for $V = 30$ and $m_v = 2500$}

\begin{figure}[h!]
	\centering
	\includegraphics{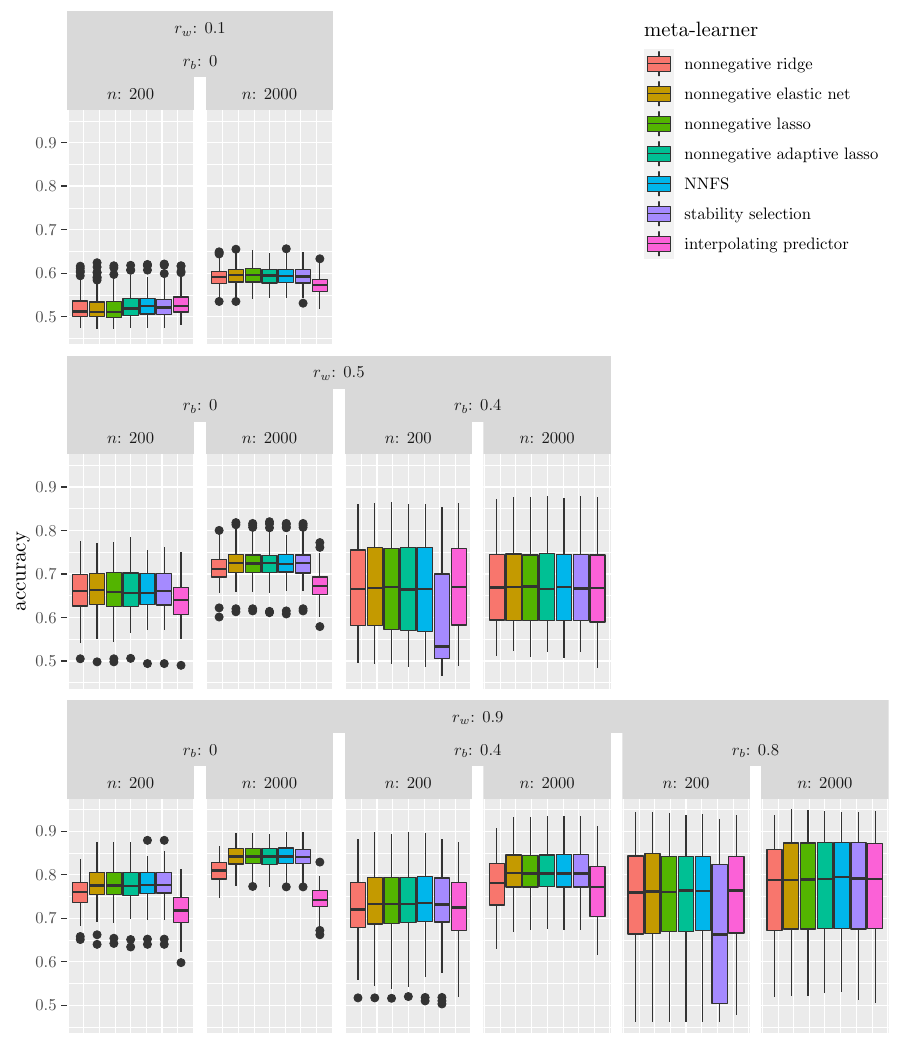}
	\caption{Boxplots of test accuracy for the different meta-learners, with 30 views and 2500 features per view. The results are shown for all combinations of the correlation between features within the same view ($\rho_w$), the correlation between features from different views ($\rho_b$), and sample size ($n$). Each plot is based on 100 replications.\label{fig:acc_v30_mv2500}}
\end{figure}

\begin{figure}[h!]
	\centering
	\includegraphics{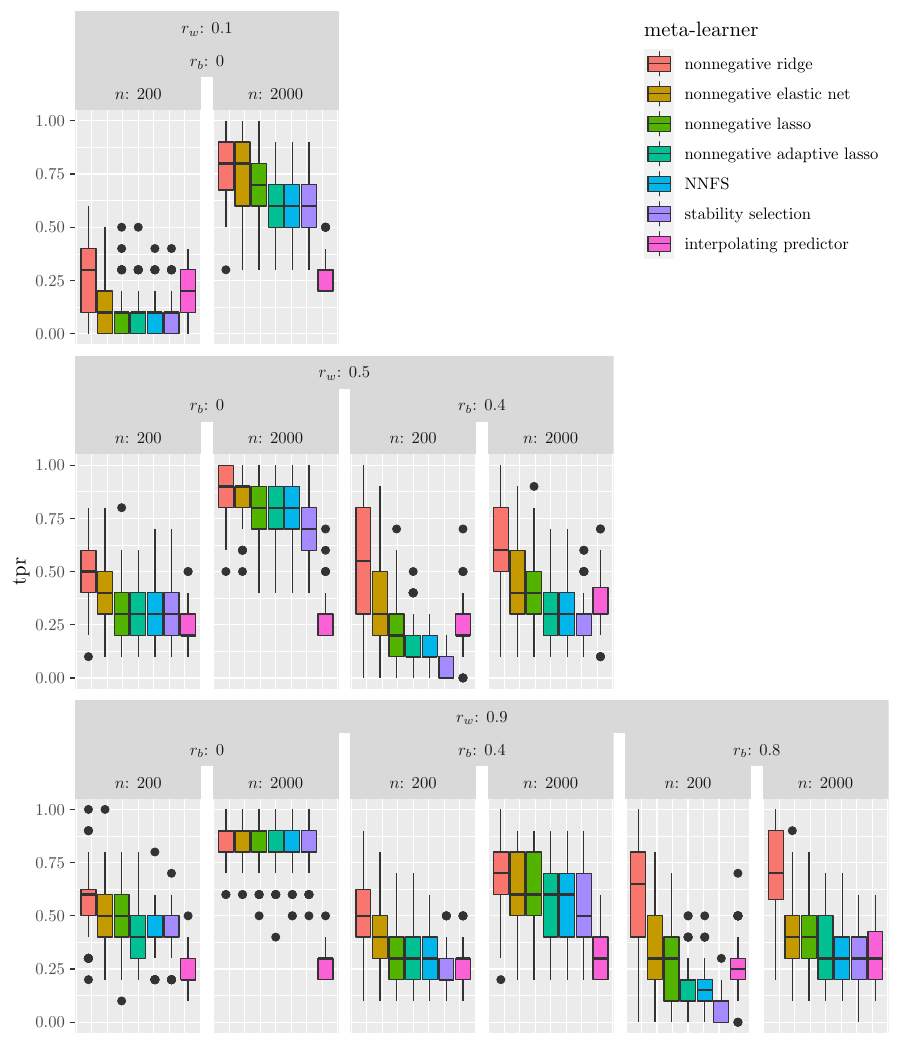}
	\caption{Boxplots of the true positive rate (TPR) for the different meta-learners, with 30 views and 2500 features per view. The results are shown for all combinations of the correlation between features within the same view ($\rho_w$), the correlation between features from different views ($\rho_b$), and sample size ($n$). Each plot is based on 100 replications.}
\end{figure}

\begin{figure}[h!]
	\centering
	\includegraphics{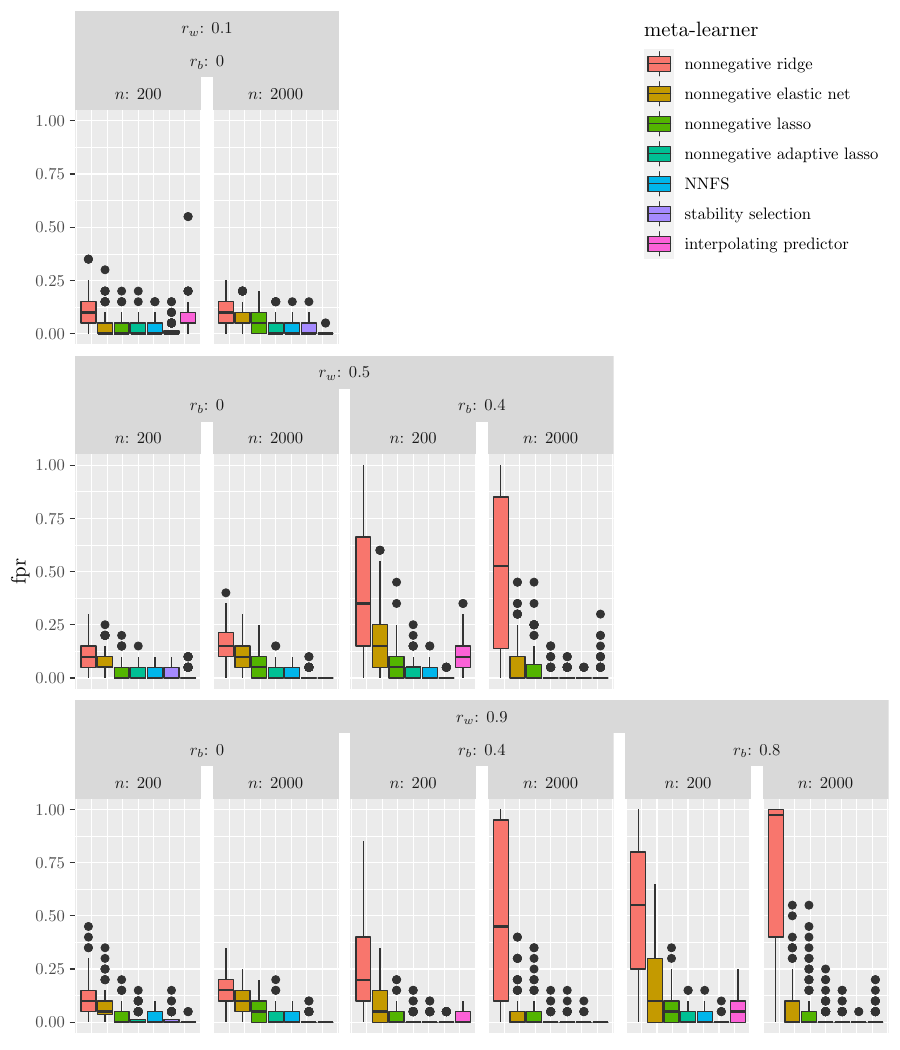}
	\caption{Boxplots of the false positive rate (FPR) for the different meta-learners, with 30 views and 2500 features per view. The results are shown for all combinations of the correlation between features within the same view ($\rho_w$), the correlation between features from different views ($\rho_b$), and sample size ($n$). Each plot is based on 100 replications.}
\end{figure}

\begin{figure}[h!]
	\centering
	\includegraphics{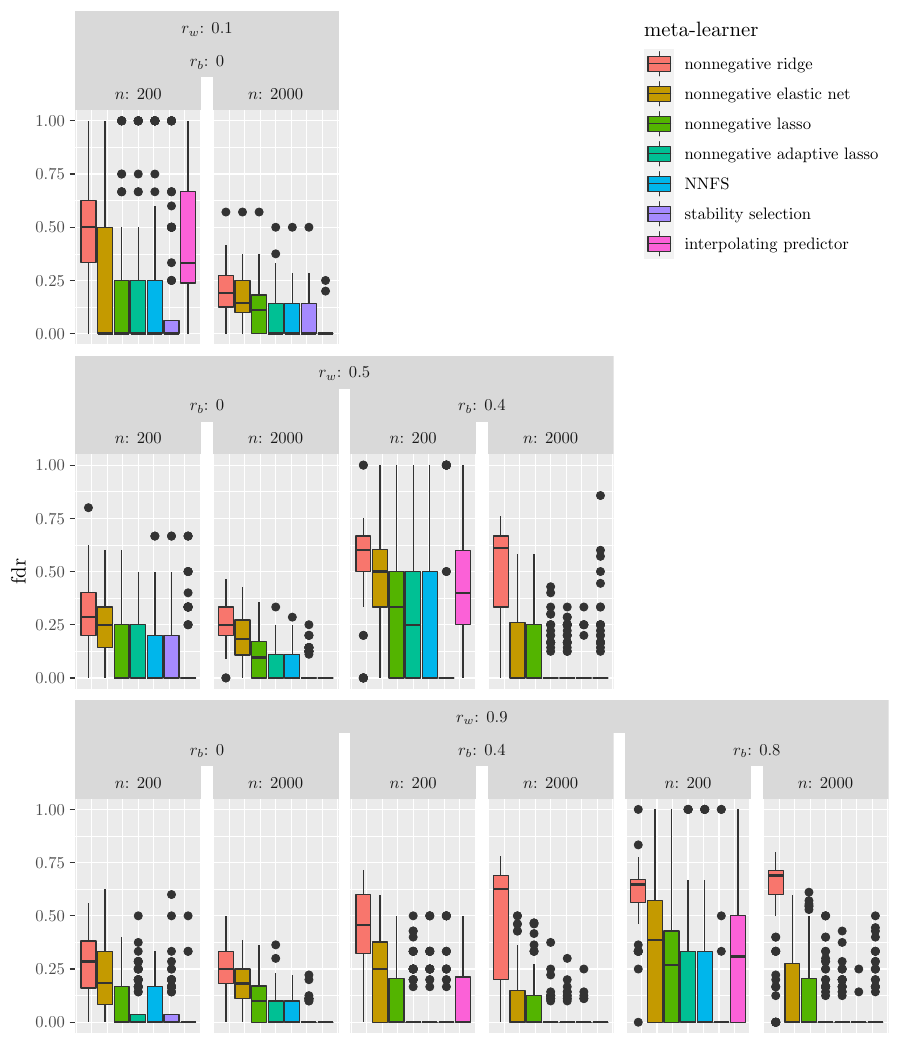}
	\caption{Boxplots of the false discovery rate (FDR) for the different meta-learners, with 30 views and 2500 features per view. The results are shown for all combinations of the correlation between features within the same view ($\rho_w$), the correlation between features from different views ($\rho_b$), and sample size ($n$). Each plot is based on 100 replications.}
\end{figure}

\FloatBarrier

\bibliographystyle{apacite}
\bibliography{Bibliography} 


\end{document}